\documentclass[researcharticle,9pt,twoside,oneauthor]{Definitions/preprint}
\usepackage{epstopdf}
\usepackage{tabularx,array}
\newcolumntype{L}{>{\raggedright\arraybackslash}X}

\runninghead{Leakage-free evaluation of residual GCNs for intrusion detection}

\Title{Leakage-free evaluation of residual graph convolutional networks for network intrusion detection}

\Author{Zahid Iqbal\textsuperscript{1,}*}
\AuthorNames{Zahid Iqbal}
\AuthorCitation{Iqbal, Z.}
\address{%
	\textsuperscript{1}\quad Department of Computer Science, Faculty of Computing and Information Technology, University of Gujrat, 50700, Punjab, Pakistan}
\corres{* Corresponding author: Zahid Iqbal}

\abstract{%
\textbf{Introduction:} Graph neural networks are increasingly proposed for network intrusion detection, yet many reported
results come from protocols in which test flows shape the graph used for training, duplicated flows sit on both sides of
the split, or a single split is reported. This study measures what a residual graph convolutional network gains over
simpler models once these sources of optimism are removed.\\[6pt]
\textbf{Materials and methods:} CICIDS2017 and NF-UNSW-NB15-v2 were cleaned with label-aware removal of exact duplicates,
which removed 69\% of the NF-UNSW-NB15-v2 flows. Flows were linked in a nearest-neighbour graph built from training flows
only, and validation and test flows were attached to their nearest training flows at prediction time. A two-layer residual
graph convolutional network was compared with eight models trained under the same fixed settings, including a
nearest-neighbour vote over the same graph, gradient-boosted trees and a random forest, over three seeds that each draw a
new split, together with ablations, a leave-one-family-out test and a feature-noise test.\\[6pt]
\textbf{Results:} On CICIDS2017 the residual network reached a binary F1 of 98.11\% (standard deviation 0.18) and on
NF-UNSW-NB15-v2 90.57\% (0.38), indistinguishable from a multilayer perceptron without the graph and from the other graph
networks. Tree ensembles were 1.48 to 1.57 points higher in F1 on CICIDS2017 and led on the area under the
precision-recall curve on both datasets; their multiclass macro-F1 on CICIDS2017 reached 93.08\% against 64.51\%, a gap that came mainly from benign flows
the residual network assigned to small attack classes. No
ablation, including transductive graph construction, changed F1 beyond the variation between seeds. Under Gaussian feature
noise with a standard deviation of 0.5, the residual network kept an F1 of 96.16\%, while XGBoost fell to 0.67\%.\\[6pt]
\textbf{Conclusions:} Under a leakage-controlled protocol, message passing over nearest-neighbour flow graphs adds no
measurable accuracy over a perceptron, and tree ensembles remain the stronger detectors on both benchmarks; the graph
models' advantage lies in robustness to feature noise.%
}


\begin{document}

\section{Introduction}

Network intrusion detection systems (NIDS) classify traffic records, usually aggregated into flows, as benign or as one
of several attack types. Most learning-based detectors treat each flow as an independent row of features. Graph neural
networks (GNNs) relax that assumption: they let a flow's prediction depend on related flows, either through the hosts the
flows connect or through the similarity of their features \cite{bilot2023}. Several recent systems report high detection scores
on public benchmarks with such models \cite{caville2022,guerra2025,ngo2025}.

Scores near the ceiling are hard to interpret on these benchmarks. Feature analysis on CICIDS2017 \cite{sharafaldin2018} has
ranked flow attributes by information gain to reduce the input dimension \cite{kurniabudi2020}, while studies of the dataset
itself report mislabelled flows and flow-construction artefacts, and corrected releases change the measured accuracy of
standard classifiers \cite{engelen2021,liu2022,lanvin2023}. A systematisation of machine learning for NIDS argues for
evaluation under conditions closer to deployment, including realistic splits and the costs of false alarms, and shows how
evaluation choices, rather than model design, can drive reported gains \cite{apruzzese2023}. On tabular intrusion data,
tree-based classifiers were not outperformed by the deep networks evaluated in a recent comparison \cite{zoppi2024}, which
motivates strong tree baselines in any study that proposes a neural detector. Feature selection combined with
gradient-boosted trees and random forests has also classified malicious activity well in other security data, for example
on the Ethereum network \cite{ehsan2024}. In monitoring applications more broadly, comparative studies of several learners
on the same data are a common basis for choosing a model \cite{iqbal2018}.

A survey groups GNN-based NIDS by the graph they build: host graphs, in which IP endpoints are nodes and flows are edges,
and flow or similarity graphs, in which flows are nodes \cite{bilot2023}. E-GraphSAGE classifies the edges of a host graph
using edge features \cite{lo2022}, and Anomal-E learns self-supervised edge embeddings on the same type of graph
\cite{caville2022}. Graph embedding \cite{xiao2020} and alternative network representations \cite{friji2023} have also been studied,
and recent self-supervised graph learning reports strong results on NetFlow-format benchmarks \cite{guerra2025}. Closest to
the present work, TKSGF connects each flow to its Top-$K$ most similar flows by attributes and applies GraphSAGE on
NF-ToN-IoT and NF-BoT-IoT \cite{ngo2025}. The NetFlow-format datasets used here and in several of these works share a common
feature set \cite{sarhan2022}.

The graph convolutional network (GCN) aggregates normalised neighbour features layer by layer \cite{kipf2017}; GraphSAGE and
the graph attention network (GAT) replace this aggregation with learned neighbour functions and attention weights
\cite{hamilton2017,velickovic2018}. Repeated aggregation smooths node representations towards each other as depth grows
\cite{liq2018}, and skip connections borrowed from residual networks \cite{he2016} are a common remedy. In intrusion detection,
E-ResSAGE and E-ResGAT add residual information to GraphSAGE and GAT by concatenation \cite{chang2024}.

Graph models also add risks of their own to the evaluation. If validation and test flows are inserted into the graph used
for training, their features shape the representations of training nodes. If exact duplicates are kept, the same flow can
sit in both the training and the test set. Finally, if only one split is reported, the differences between models may be
smaller than the variation between splits.

This paper evaluates a residual graph convolutional network (R-GCN) under a protocol designed to remove these sources of
optimism. Flows are nodes of a $k$-nearest-neighbour ($k$NN) graph in feature space. The graph used for training contains
training flows only, and each validation or test flow is attached to its nearest training flows at prediction time, so its
prediction depends on its own features and on training data alone. Exact duplicate flows are removed before splitting,
every preprocessing statistic is fitted on training flows, and each of three seeds draws a new split. The R-GCN adds the
previous hidden layer through an identity skip, whereas E-ResSAGE and E-ResGAT concatenate \cite{chang2024}. Its $k$NN
graph also differs from TKSGF, which builds a separate similarity graph inside each partition and so links test flows to
one another \cite{ngo2025}; here a validation or test flow connects only to training flows. Table~\ref{tab:closest} compares these studies by
graph type, the edges given to test flows, the split and the number of repeated runs. The contribution lies in this
protocol and in the evidence it yields: what the residual GCN gains over a vote of its own neighbours and over tree
ensembles once these sources of optimism are removed, measured over resampled splits on two datasets.

\begin{table}[!htbp]
\caption{Graph-based intrusion detection studies closest to the present work, as described in the cited papers (--: not stated).}
\label{tab:closest}
\tablesize{\small}\renewcommand{\arraystretch}{1.3}
\setlength{\tabcolsep}{3pt}
\begin{tabularx}{\textwidth}{>{\raggedright\arraybackslash}p{2.3cm}>{\hsize=0.75\hsize}L>{\hsize=1.25\hsize}L>{\hsize=1.0\hsize}Ll>{\hsize=0.7\hsize}L>{\hsize=1.3\hsize}L}
\toprule
\textbf{Study} & \textbf{Nodes; edges} & \textbf{Model; learning} & \textbf{Edges of test flows} & \textbf{Split} & \textbf{Repeated runs} & \textbf{Datasets} \\
\midrule
E-GraphSAGE \cite{lo2022} & hosts; flows & E-GraphSAGE; supervised & -- & 70/30 & -- & BoT-IoT, ToN-IoT, NF-BoT-IoT, NF-ToN-IoT \\
\arrayrulecolor{black!25}\midrule[0.3pt]\arrayrulecolor{black}
Anomal-E \cite{caville2022} & hosts; flows & E-GraphSAGE + DGI; self-supervised & -- & -- & -- & NF-UNSW-NB15-v2, NF-CSE-CIC-IDS2018-v2 \\
\arrayrulecolor{black!25}\midrule[0.3pt]\arrayrulecolor{black}
E-ResSAGE, E-ResGAT \cite{chang2024} & endpoints; flows & GraphSAGE, GAT + feature concatenation; supervised & -- & 50/20/30 & -- & UNSW-NB15, CIC-DarkNet, CSE-CIC-IDS, ToN-IoT \\
\arrayrulecolor{black!25}\midrule[0.3pt]\arrayrulecolor{black}
TKSGF \cite{ngo2025} & flows; Top-$K$ cosine & GraphSAGE (GCN, GAT); supervised & among test flows (graph per partition) & 60/20/20, 70/15/15 & $\geq$ 5 & NF-ToN-IoT, NF-BoT-IoT \\
\arrayrulecolor{black!25}\midrule[0.3pt]\arrayrulecolor{black}
GraphIDS \cite{guerra2025} & hosts; flows & E-GraphSAGE + masked autoencoder; benign-only & among test flows (graph per partition) & 80/10/10 & several seeds, mean $\pm$ SD & NF-UNSW-NB15, NF-CSE-CIC-IDS2018 (v2, v3) \\
\arrayrulecolor{black!25}\midrule[0.3pt]\arrayrulecolor{black}
\textbf{This study} & flows; $k$NN, $k=10$ & R-GCN and 8 baselines; supervised & to training flows only & 60/20/20 & 3 seeds, new split each & CICIDS2017, NF-UNSW-NB15-v2 (de-duplicated) \\
\bottomrule
\end{tabularx}
\end{table}

The main contributions of this paper are as follows:
\begin{enumerate}
\item An inductive evaluation protocol for flow-graph intrusion detection, in which the $k$NN graph, the feature scaling
and the model are learned from training flows only, validation and test flows link only to training flows, exact
duplicates are removed before splitting, and each of three seeds draws a new split (Section~\ref{sec:data} and
Figure~\ref{fig:pipeline}). On NF-UNSW-NB15-v2, de-duplication alone removed 69\% of the flows.
\item A same-budget comparison of a residual graph convolutional network with eight models on CICIDS2017 and
NF-UNSW-NB15-v2, including a $k$NN vote over the same neighbours, three tree ensembles and a host-graph model, for
binary and multiclass detection with paired per-seed tests (Sections~\ref{sec:baselines} and~\ref{sec:protocol}).
\item Evidence that, under this protocol, the graph networks match a multilayer perceptron while tree ensembles remain
the strongest detectors, with a binary F1 1.48 to 1.57 points higher on CICIDS2017
(Tables~\ref{tab:bin_cic}--\ref{tab:paired} and Figure~\ref{fig:forest}).
\item A per-class diagnosis that traces the multiclass gap of the graph networks to false alarms on small attack
classes: the residual network's precision on Bot flows was 3.36\%, against 55.49\% for XGBoost
(Figures~\ref{fig:perclass} and~\ref{fig:confusion}).
\item An ablation of neighbourhood size, depth, the residual connection, the loss, normalisation, the port feature and
transductive graph construction, none of which changed F1 beyond the variation between seeds (Table~\ref{tab:abl} and
Figure~\ref{fig:kdepth}).
\item Two stress tests that show where graph models help: under Gaussian feature noise with a standard deviation of 0.5,
the residual network kept an F1 of 96.16\% while XGBoost fell to 0.67\%, and it recovered about half of a DoS family
withheld from training (Tables~\ref{tab:lofo} and~\ref{tab:noise}, Figures~\ref{fig:lofo} and~\ref{fig:noise}).
\end{enumerate}

\section{Materials and methods}
\begin{figure}[!htbp]
\centering
\includegraphics[width=\textwidth]{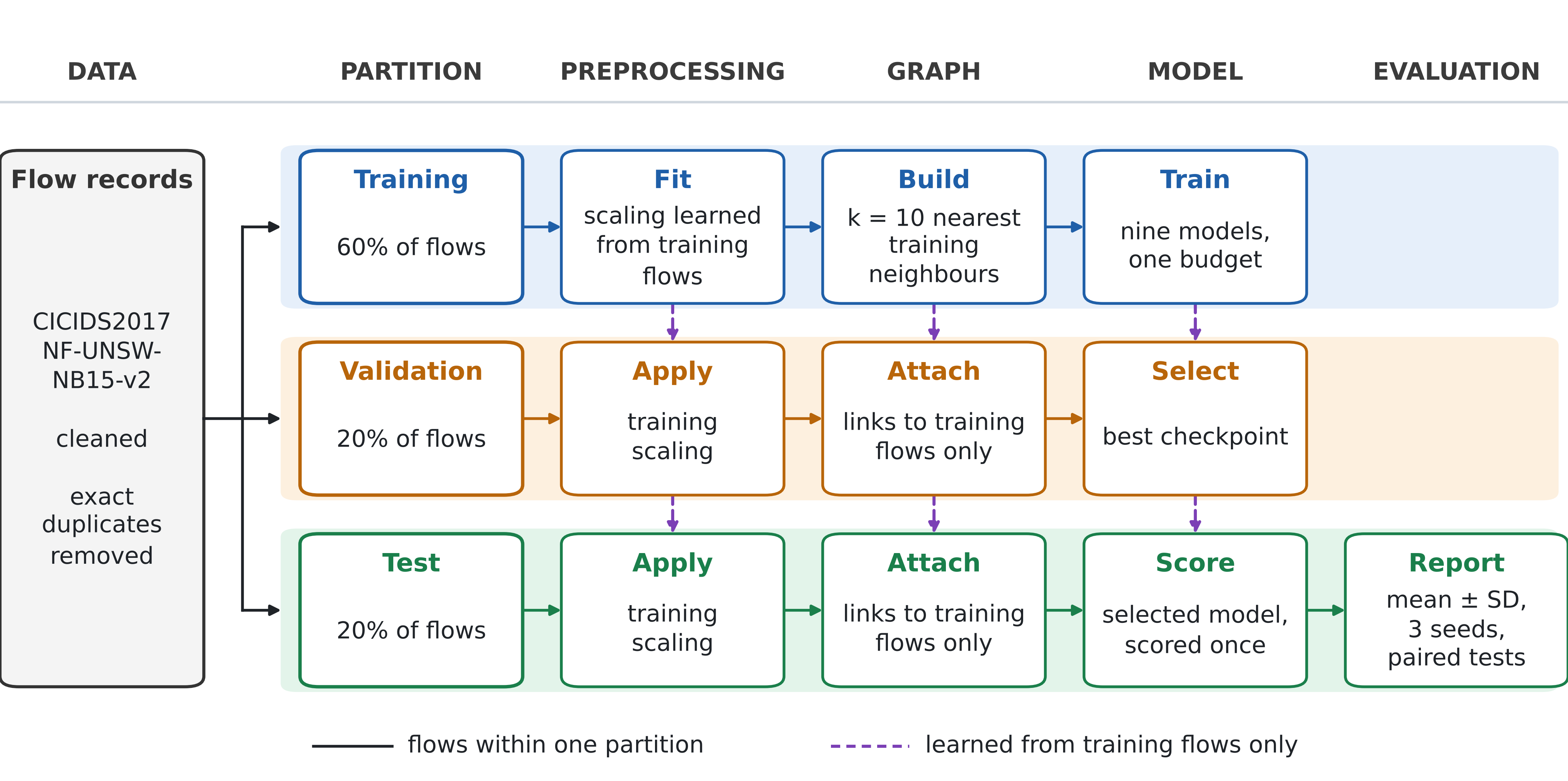}
\caption{Evaluation protocol. The flow records are cleaned and exact duplicates are removed before a stratified split
forms the training, validation and test partitions (one lane each). Everything learned from data (the scaling, the
neighbours of the $k$NN graph and the model) comes from training flows and is passed down to the other lanes (dashed
arrows); validation and test flows link only to training flows. Seeds 0, 1 and 2 repeat the protocol with new splits.}
\label{fig:pipeline}
\end{figure}

Figure~\ref{fig:pipeline} summarises the procedure from raw flow records to the reported scores.

\subsection{Datasets, cleaning and splits}\label{sec:data}
\textbf{CICIDS2017.} The eight labelled flow files (one per capture session) published with the dataset
\cite{sharafaldin2018} were used. Identifier columns (flow identifier, source and destination addresses, source
port, timestamp) and one duplicated header column were removed. Labels were mapped to eight families: Benign, DoS, DDoS,
PortScan, BruteForce (FTP and SSH password guessing), Web, Bot and Rare (Infiltration, Heartbleed and any other label).

\textbf{NF-UNSW-NB15-v2.} The NetFlow version 2 release of UNSW-NB15 \cite{moustafa2015,sarhan2022} provides 43 NetFlow
features per flow. The source port, the DNS query identifier and the two label columns were removed; the IP addresses were kept aside as host
identifiers for the host-graph baseline and excluded from the features. Attack classes with fewer than 500 flows were merged
into a Rare class (here, Worms).

\textbf{Cleaning.} For both datasets, infinite values were treated as missing, rows with missing values were removed, and
exact duplicate rows were removed when they had the same features and the same class (label-aware de-duplication), so that
no flow can occur in both the training and the test partition. Constant columns were dropped. CICIDS2017 went from
2,830,743 to 2,520,798 flows with 69 columns, and NF-UNSW-NB15-v2 from 2,390,275 to 731,810 flows with 39 columns; the
destination-port column is excluded at training time, as described below, which leaves 68 and 38 model inputs.
Tables~\ref{tab:data_cic} and~\ref{tab:data_nf} list the class counts. Each test partition (20\% of the flows) holds 504,160 flows for
CICIDS2017 and 146,362 for NF-UNSW-NB15-v2.

\begin{table}[!htbp]
\begin{minipage}[t]{0.48\textwidth}
\centering
\caption{Class distribution of CICIDS2017 after cleaning.}
\label{tab:data_cic}
\tablesize{\small}\renewcommand{\arraystretch}{1.3}
\setlength{\tabcolsep}{4pt}
\begin{tabularx}{\linewidth}{Lrr}
\toprule
\textbf{Class} & \textbf{Flows} & \textbf{Share (\%)} \\
\midrule
Benign & 2,095,057 & 83.111 \\
\arrayrulecolor{black!25}\midrule[0.3pt]\arrayrulecolor{black}
DoS & 193,745 & 7.686 \\
\arrayrulecolor{black!25}\midrule[0.3pt]\arrayrulecolor{black}
DDoS & 128,014 & 5.078 \\
\arrayrulecolor{black!25}\midrule[0.3pt]\arrayrulecolor{black}
PortScan & 90,694 & 3.598 \\
\arrayrulecolor{black!25}\midrule[0.3pt]\arrayrulecolor{black}
BruteForce & 9,150 & 0.363 \\
\arrayrulecolor{black!25}\midrule[0.3pt]\arrayrulecolor{black}
Web & 2,143 & 0.085 \\
\arrayrulecolor{black!25}\midrule[0.3pt]\arrayrulecolor{black}
Bot & 1,948 & 0.077 \\
\arrayrulecolor{black!25}\midrule[0.3pt]\arrayrulecolor{black}
Rare & 47 & 0.002 \\
\midrule
Total & 2,520,798 & 100.000 \\
\bottomrule
\end{tabularx}
\end{minipage}
\hfill
\begin{minipage}[t]{0.48\textwidth}
\centering
\caption{Class distribution of NF-UNSW-NB15-v2 after cleaning.}
\label{tab:data_nf}
\tablesize{\small}\renewcommand{\arraystretch}{1.3}
\setlength{\tabcolsep}{4pt}
\begin{tabularx}{\linewidth}{Lrr}
\toprule
\textbf{Class} & \textbf{Flows} & \textbf{Share (\%)} \\
\midrule
Benign & 690,559 & 94.363 \\
\arrayrulecolor{black!25}\midrule[0.3pt]\arrayrulecolor{black}
Analysis & 762 & 0.104 \\
\arrayrulecolor{black!25}\midrule[0.3pt]\arrayrulecolor{black}
Backdoor & 717 & 0.098 \\
\arrayrulecolor{black!25}\midrule[0.3pt]\arrayrulecolor{black}
DoS & 3,813 & 0.521 \\
\arrayrulecolor{black!25}\midrule[0.3pt]\arrayrulecolor{black}
Exploits & 23,565 & 3.220 \\
\arrayrulecolor{black!25}\midrule[0.3pt]\arrayrulecolor{black}
Fuzzers & 7,405 & 1.012 \\
\arrayrulecolor{black!25}\midrule[0.3pt]\arrayrulecolor{black}
Generic & 2,372 & 0.324 \\
\arrayrulecolor{black!25}\midrule[0.3pt]\arrayrulecolor{black}
Reconnaissance & 1,121 & 0.153 \\
\arrayrulecolor{black!25}\midrule[0.3pt]\arrayrulecolor{black}
Shellcode & 1,427 & 0.195 \\
\arrayrulecolor{black!25}\midrule[0.3pt]\arrayrulecolor{black}
Rare & 69 & 0.009 \\
\midrule
Total & 731,810 & 100.000 \\
\bottomrule
\end{tabularx}
\end{minipage}
\end{table}

\label{sec:prep}%
Each seed $s\in\{0,1,2\}$ draws a new stratified split into training, validation and test partitions (60\%, 20\% and 20\%
of flows); the same seed initialises the network weights and the tree ensembles. All preprocessing is fitted on the
training partition and then applied unchanged to the other two. The destination port is excluded from the features by
default, because a port number can act as a shortcut to the label; an ablation adds it back. A column $j$ receives a signed
logarithmic transform when it is non-negative on all training flows and its 99th percentile on the training flows
exceeds 10:
\begin{equation}
  \tilde{x}_{j} = \operatorname{sign}(x_j)\,\log\!\left(1+|x_j|\right).
\end{equation}
On training flows this equals $\log(1+x_j)$; the signed form keeps the transform finite for negative values that occur in
some validation or test flows of CICIDS2017. Every column is then standardised with the training mean and standard
deviation and clipped to $[-10,10]$.

\subsection{Inductive $k$NN flow graph and residual network}\label{sec:graph}
\begin{figure}[!htbp]
\centering
\includegraphics[width=0.72\textwidth]{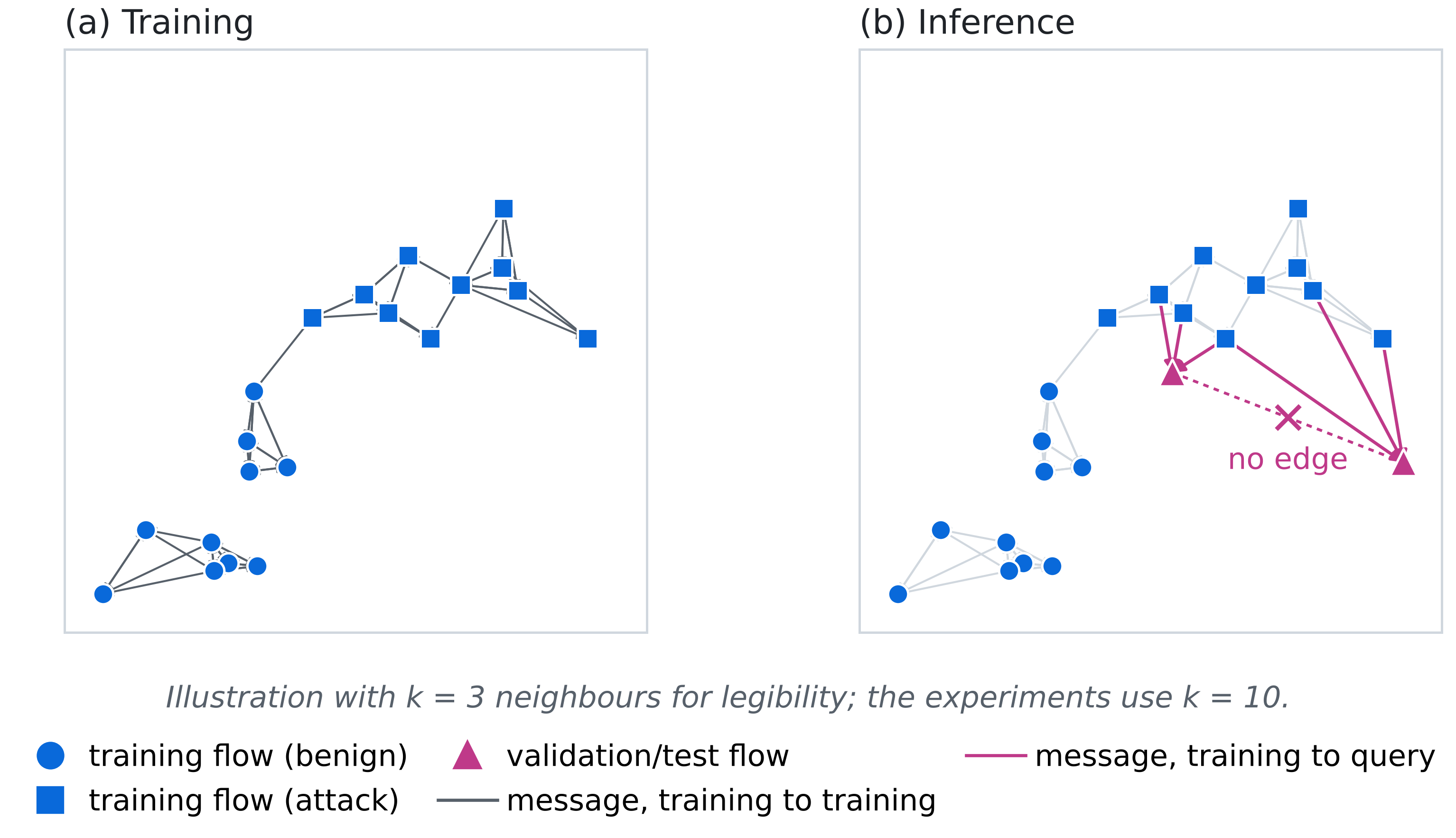}
\caption{Inductive $k$NN flow graph (schematic with $k=3$; the experiments use $k=10$). (a) During training, each
training flow receives messages from its nearest training flows. (b) A validation or test flow receives messages from its
nearest training flows only; query flows are never linked to each other and never send messages to training flows.}
\label{fig:graph}
\end{figure}

Let $\mathcal{T}$ be the set of training flows and $\mathbf{x}_i$ the preprocessed feature vector of flow $i$. For a training
flow $i$, $\mathcal{N}_k(i)$ contains the $k$ training flows closest to $\mathbf{x}_i$ in Euclidean distance, excluding $i$
itself. For a validation or test flow $q$, $\mathcal{N}_k(q)$ contains the $k$ training flows closest to $\mathbf{x}_q$. Messages
flow only from training flows to their neighbours, so the graph seen during training contains training flows only, and a
query flow's prediction depends on its own features and on the training data. No edge joins two query flows, and no
training flow receives a message from a query flow. Neighbours are found by exact search, with
$k=10$ by default. Figure~\ref{fig:graph} illustrates the construction.

\begin{figure}[!htbp]
\centering
\includegraphics[width=\textwidth]{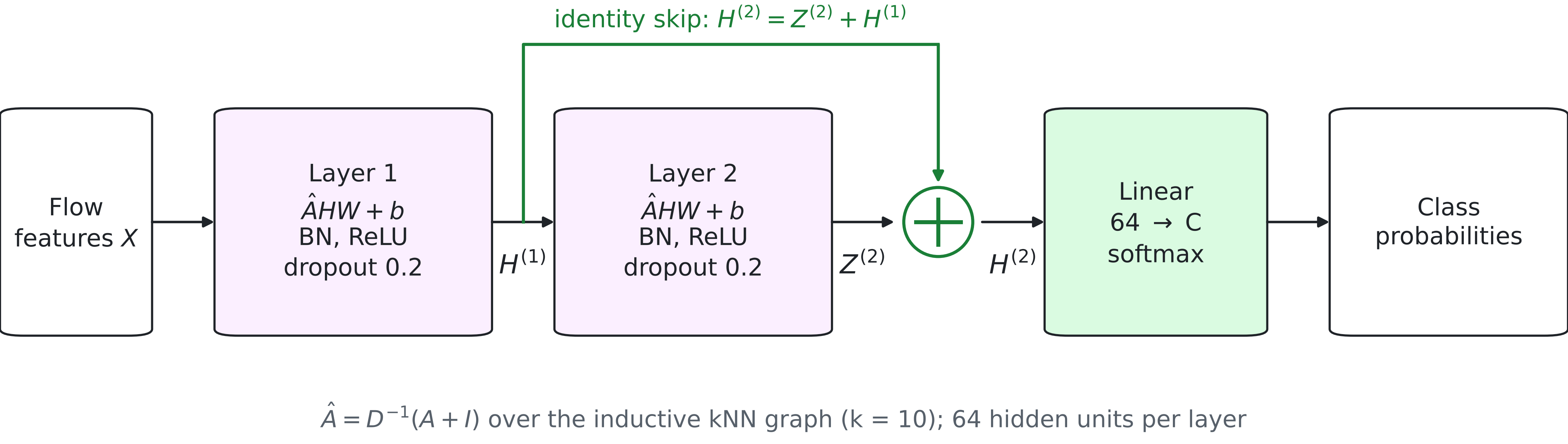}
\caption{Two-layer residual GCN. Each layer applies mean aggregation $\hat{A}=D^{-1}(A+I)$ over the inductive $k$NN graph,
a linear map, batch normalisation, ReLU and dropout; the output of the first layer is added to that of the second before
the classifier.}
\label{fig:rgcn}
\end{figure}

\label{sec:rgcn}%
Each layer averages a node's transformed features with those of its neighbours, using the row-normalised operator
\begin{equation}
  \hat{A} = D^{-1}(A+I), \qquad \left(\hat{A}\,Z\right)_i = \frac{1}{|\mathcal{N}_k(i)|+1}\Big(\mathbf{z}_i + \sum_{j\in\mathcal{N}_k(i)} \mathbf{z}_j\Big),
\end{equation}
where $A$ is the adjacency matrix of the $k$NN graph (with $A_{ij}=1$ when $j\in\mathcal{N}_k(i)$) and $D$ is the diagonal
degree matrix of $A+I$. This is a mean aggregation over a node and its neighbours; it differs from the symmetric
normalisation $\tilde{D}^{-1/2}\tilde{A}\tilde{D}^{-1/2}$ of the original GCN \cite{kipf2017} because the $k$NN graph is
directed. Layer $l$ computes
\begin{equation}
  Z^{(l)} = \operatorname{Dropout}\!\Big(\operatorname{ReLU}\big(\operatorname{BN}(\hat{A}\,H^{(l-1)}W^{(l)} + \mathbf{b}^{(l)})\big)\Big),
\end{equation}
with $H^{(0)}=X$, batch normalisation (BN) \cite{ioffe2015} and dropout probability 0.2. The residual connection adds the previous hidden
representation from the second layer onwards:
\begin{equation}
  H^{(1)} = Z^{(1)}, \qquad H^{(l)} = Z^{(l)} + H^{(l-1)} \quad (l\ge 2).
\end{equation}
With two layers this is $H^{(2)} = Z^{(2)} + Z^{(1)}$. A linear layer maps $H^{(L)}$ to class scores, followed by the
softmax. Both hidden layers have 64 units (Figure~\ref{fig:rgcn}).

\textbf{Loss.} Classes are weighted by $w_c=\min\!\left(N/(C\,n_c),\,100\right)$, where $N$ is the number of training flows,
$C$ the number of classes and $n_c$ the training count of class $c$. The weighted cross-entropy is
\begin{equation}
  \mathcal{L} = -\frac{\sum_{i\in\mathcal{T}} w_{y_i}\log p_{i,y_i}}{\sum_{i\in\mathcal{T}} w_{y_i}},
\end{equation}
where $p_{i,c}$ is the softmax probability of class $c$ for flow $i$. The ablation also tests the unweighted cross-entropy
and a focal loss with $\gamma=2$ and the same class weights.

\textbf{Training and model selection.} Networks were trained full-batch with Adam \cite{kingma2015} (learning rate 0.01) for 200 epochs.
The validation score was computed at epoch 1, every fifth epoch and the last epoch, and the weights with the best
validation score were restored. The validation score is the area under the precision-recall curve (PR-AUC) for binary
detection and the macro-averaged F1 for multiclass detection. No hyperparameter was tuned on the test partition, and the
same fixed settings were used for every seed.

\subsection{Baselines}\label{sec:baselines}
All neural models share the layer structure, width, loss, optimiser, epoch budget and model-selection rule of the R-GCN
and differ only in how neighbours are aggregated: a GCN without the residual connection, GraphSAGE with mean aggregation
\cite{hamilton2017}, a GAT with four heads \cite{velickovic2018}, and a multilayer perceptron (MLP) that ignores the graph.
The $k$NN vote predicts the class shares among the same $k$ training neighbours used by the graph models,
\begin{equation}
  p_c(q)=\frac{1}{k}\sum_{j\in\mathcal{N}_k(q)}\mathbb{1}[y_j=c],
\end{equation}
which tests whether message passing adds anything beyond the labels of the neighbours. The tabular baselines are XGBoost
\cite{chen2016} (400 trees, depth 8, learning rate 0.1, row and column subsampling 0.8, early stopping after 20 rounds on the
validation partition), LightGBM \cite{ke2017} (400 trees, 63 leaves, depth 8, learning rate 0.1, the same subsampling and
early stopping) and a random forest \cite{breiman2001} (100 trees, 30\% of the training flows per tree, class weights balanced
within each bootstrap sample). XGBoost and LightGBM use the class weights $w_c$ as sample weights. E-GraphSAGE \cite{lo2022} is run on a host
graph in which IP addresses are nodes and flows are edges with their features. It uses the shared settings of
Table~\ref{tab:hyper} (64 hidden units, learning rate 0.01) rather than the 128 units and learning rate 0.001 of the original
study \cite{lo2022}, so that every neural model has the same budget. Because each partition forms its own host graph, its
results are reported separately from the $k$NN-graph models. Table~\ref{tab:hyper} lists every
setting.

\begin{table}[!htbp]
\caption{Settings of all models. The same values were used for every seed and for both datasets.}
\label{tab:hyper}
\tablesize{\small}\renewcommand{\arraystretch}{1.3}
\setlength{\tabcolsep}{3pt}
\begin{tabularx}{\textwidth}{lL}
\toprule
\textbf{Setting} & \textbf{Value} \\
\midrule
\multicolumn{2}{l}{\textit{Neural models (shared)}} \\
Graph & $k$NN of flows, $k=10$, Euclidean distance, inductive (training flows only) \\
\arrayrulecolor{black!25}\midrule[0.3pt]\arrayrulecolor{black}
Layers / hidden units & 2 / 64 \\
\arrayrulecolor{black!25}\midrule[0.3pt]\arrayrulecolor{black}
Normalisation / dropout & batch normalisation; dropout 0.2 \\
\arrayrulecolor{black!25}\midrule[0.3pt]\arrayrulecolor{black}
Residual (R-GCN only) & $H^{(l)}=Z^{(l)}+H^{(l-1)}$ for $l\ge 2$ \\
\arrayrulecolor{black!25}\midrule[0.3pt]\arrayrulecolor{black}
GAT heads / GraphSAGE & 4 heads / mean aggregation \\
\arrayrulecolor{black!25}\midrule[0.3pt]\arrayrulecolor{black}
Loss & weighted cross-entropy, $w_c=\min(N/(C n_c),100)$ \\
\arrayrulecolor{black!25}\midrule[0.3pt]\arrayrulecolor{black}
Optimiser & Adam, learning rate 0.01, full batch, 200 epochs \\
\arrayrulecolor{black!25}\midrule[0.3pt]\arrayrulecolor{black}
Model selection & validation score at epoch 1, every 5th epoch and epoch 200; best weights restored (PR-AUC binary, macro-F1 multiclass) \\
\arrayrulecolor{black!25}\midrule[0.3pt]\arrayrulecolor{black}
$k$NN vote & class shares of the same $k=10$ training neighbours; no training \\
\midrule
\multicolumn{2}{l}{\textit{Tree ensembles}} \\
XGBoost & 400 trees, depth 8, learning rate 0.1, row/column subsampling 0.8/0.8, histogram tree method, early stopping 20 rounds, sample weights $w_c$ \\
\arrayrulecolor{black!25}\midrule[0.3pt]\arrayrulecolor{black}
LightGBM & 400 trees, 63 leaves, depth 8, learning rate 0.1, row/column subsampling 0.8/0.8, early stopping 20 rounds, sample weights $w_c$ \\
\arrayrulecolor{black!25}\midrule[0.3pt]\arrayrulecolor{black}
Random forest & 100 trees, 30\% of training flows per tree, class weights balanced per bootstrap sample \\
\midrule
\multicolumn{2}{l}{\textit{Host-graph baseline}} \\
E-GraphSAGE & 2 layers, 64 hidden units, dropout 0.2, Adam 0.01, 200 epochs, same loss and selection rule \\
\bottomrule
\end{tabularx}
\end{table}

\subsection{Evaluation protocol, metrics and stress tests}\label{sec:protocol}
Binary detection treats every attack class as positive; multiclass detection uses all classes of a dataset. For binary
detection we report F1, the Matthews correlation coefficient (MCC), PR-AUC and the false-positive rate (FPR) at the
decision threshold 0.5, together with the recall at fixed false-positive rates of 0.1\% and 1\%, which still separates
models when F1 is near its ceiling. For multiclass detection we report accuracy, macro-averaged F1 over the classes present
in the test partition, weighted F1 and per-class F1. Every metric is the mean and standard deviation over the three
seeds. Models are compared through paired per-seed differences with a 95\% confidence interval from the $t$ distribution
and a paired $t$-test whose $p$-values are Holm-corrected within each dataset and metric; with three seeds, a difference is
treated as real only when its interval excludes zero.

The ablation changes one setting of the R-GCN at a time on CICIDS2017: $k\in\{5,20,50\}$; one layer; four layers without a
skip, with the residual connection, and with an additional input projection; unweighted and focal losses; no batch
normalisation; the destination port added back; and transductive construction, in which one $k$NN graph is built over
training, validation and test features and batch-normalisation statistics are computed over all nodes. The transductive
variant uses test features but never test labels.

\label{sec:lofo}%
In the leave-one-family-out test, all flows of one CICIDS2017 attack family (DoS, DDoS, PortScan, BruteForce, Web or Bot)
are removed from the training and validation partitions while the test partition is unchanged. The binary detector is
then scored by its recall on the held-out family, its recall on the families seen in training and its FPR. In the noise
test, Gaussian noise with standard deviation 0.1, 0.25 and 0.5 (in units of the standardised features) is added to up to
100,000 test flows, values are clipped to $[-10,10]$, and graph models re-link each noisy flow to its nearest training
flows.

\subsection{Implementation}\label{sec:impl}
All models were implemented in Python 3.11 with PyTorch 2.14.1, PyTorch Geometric 2.8.1, scikit-learn 1.9.0, XGBoost
3.2.0 and LightGBM 4.7.0, and trained on an NVIDIA A100 80 GB GPU. The full set of 216 runs (Sections~\ref{sec:data}
to~\ref{sec:protocol}) was fixed before the experiments started, and every run is reported. 

\section{Results}
This section first reports binary detection on both datasets, together with the paired differences that show whether
the residual network differs from each competitor, and then multiclass detection. The ablations follow, and the last
subsection reports two stress tests: attack families withheld from training and Gaussian noise on the test features. All
values are means over three seeds, each with its own split.

\subsection{Binary detection}
\begin{figure}[!htbp]
\centering
\includegraphics[width=\textwidth]{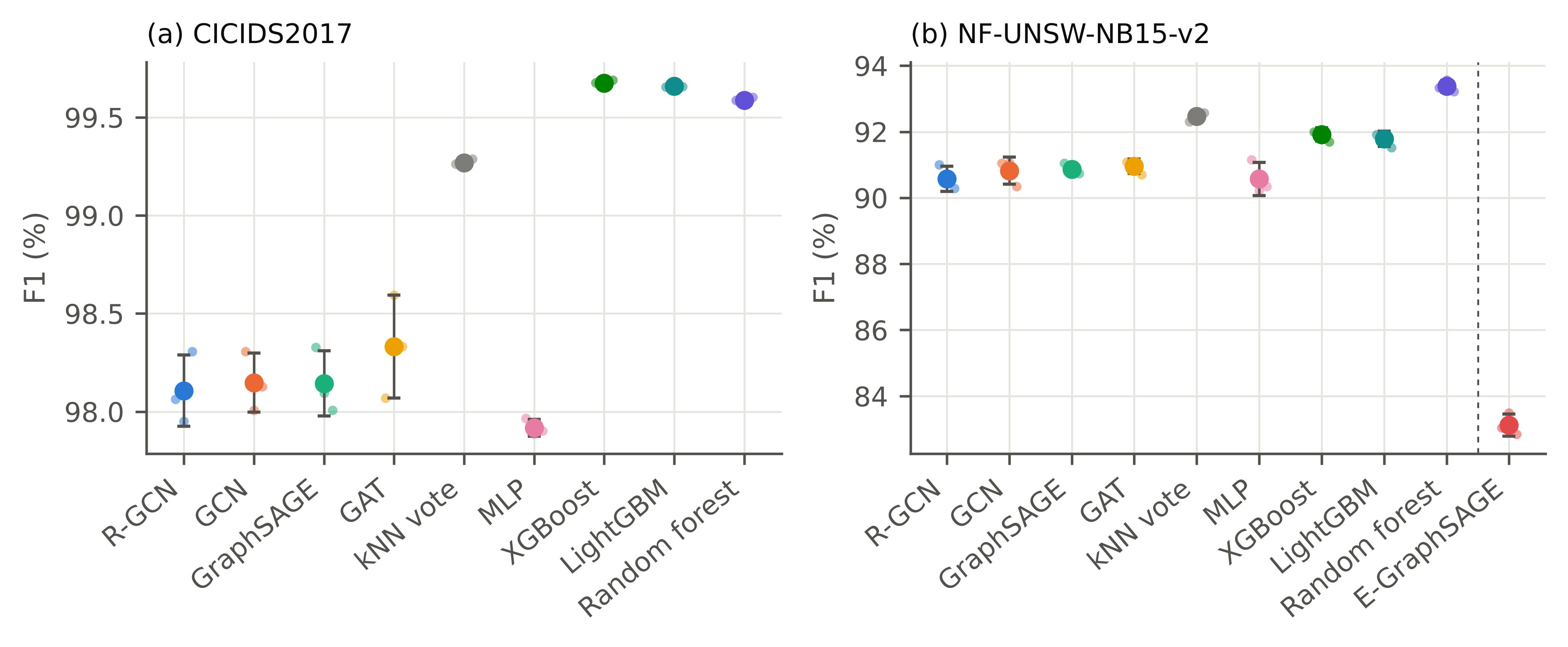}
\caption{Binary F1 per model on the test partitions. Small points: individual seeds; large points: mean $\pm$ SD over three
seeds. The F1 axis does not start at zero. E-GraphSAGE (right of the dashed line) uses a host graph and is reported
separately.}
\label{fig:f1}
\end{figure}

Tables~\ref{tab:bin_cic} and~\ref{tab:bin_nf} report binary detection on the test partitions, and
Figure~\ref{fig:f1} shows the F1 of every seed. On CICIDS2017 the five neural
models scored between 97.92\% (MLP) and 98.33\% (GAT) in F1, with the R-GCN at 98.11\%. The three tree ensembles scored
between 99.59\% and 99.67\%, and the $k$NN vote over the same neighbours reached 99.27\%. The ensembles also had the lowest
false-positive rates (0.08\% to 0.13\% of benign flows, against 0.64\% for the R-GCN) and the highest recall at a
false-positive rate of 0.1\% (99.70\% to 99.88\%, against 97.39\%). On NF-UNSW-NB15-v2 all models scored lower. The neural
models lay between 90.57\% and 90.96\% in F1, the random forest reached 93.38\%, and XGBoost had the highest PR-AUC
(96.94\%, against 95.11\% for the R-GCN) and the highest recall at a false-positive rate of 0.1\% (56.95\%, against
43.62\%). The host-graph model E-GraphSAGE reached 83.12\% in F1 and 78.23\% in PR-AUC.

\begin{table}[!htbp]
\caption{Binary detection on the CICIDS2017 test partitions (\%, MCC without unit; mean $\pm$ SD over three seeds). Recall at 0.1\% or 1\% FPR is the highest recall at a false-positive rate not above that value; bold marks the best mean.}
\label{tab:bin_cic}
\tablesize{\small}\renewcommand{\arraystretch}{1.3}
\setlength{\tabcolsep}{3pt}
\begin{tabularx}{\textwidth}{lCCCCCC}
\toprule
\textbf{Model} & \textbf{F1} & \textbf{MCC} & \textbf{PR-AUC} & \textbf{FPR} & \textbf{Recall at 0.1\% FPR} & \textbf{Recall at 1\% FPR} \\
\midrule
R-GCN & 98.11 $\pm$ 0.18 & 0.977 $\pm$ 0.002 & 99.86 $\pm$ 0.01 & 0.64 $\pm$ 0.07 & 97.39 $\pm$ 0.14 & 99.52 $\pm$ 0.04 \\
\arrayrulecolor{black!25}\midrule[0.3pt]\arrayrulecolor{black}
GCN & 98.15 $\pm$ 0.15 & 0.978 $\pm$ 0.002 & 99.86 $\pm$ 0.01 & 0.62 $\pm$ 0.06 & 97.67 $\pm$ 0.18 & 99.53 $\pm$ 0.03 \\
\arrayrulecolor{black!25}\midrule[0.3pt]\arrayrulecolor{black}
GraphSAGE & 98.14 $\pm$ 0.17 & 0.978 $\pm$ 0.002 & 99.86 $\pm$ 0.00 & 0.62 $\pm$ 0.09 & 97.40 $\pm$ 0.29 & 99.49 $\pm$ 0.01 \\
\arrayrulecolor{black!25}\midrule[0.3pt]\arrayrulecolor{black}
GAT & 98.33 $\pm$ 0.26 & 0.980 $\pm$ 0.003 & 99.85 $\pm$ 0.02 & 0.51 $\pm$ 0.17 & 97.29 $\pm$ 0.85 & 99.52 $\pm$ 0.05 \\
\arrayrulecolor{black!25}\midrule[0.3pt]\arrayrulecolor{black}
$k$NN vote & 99.27 $\pm$ 0.02 & 0.991 $\pm$ 0.000 & 99.55 $\pm$ 0.01 & 0.15 $\pm$ 0.01 & 98.71 $\pm$ 0.07 & 99.81 $\pm$ 0.01 \\
\arrayrulecolor{black!25}\midrule[0.3pt]\arrayrulecolor{black}
MLP & 97.92 $\pm$ 0.04 & 0.975 $\pm$ 0.001 & 99.84 $\pm$ 0.01 & 0.71 $\pm$ 0.02 & 97.11 $\pm$ 0.28 & 99.43 $\pm$ 0.04 \\
\arrayrulecolor{black!25}\midrule[0.3pt]\arrayrulecolor{black}
XGBoost & \textbf{99.67 $\pm$ 0.02} & \textbf{0.996 $\pm$ 0.000} & \textbf{99.99 $\pm$ 0.00} & 0.12 $\pm$ 0.01 & \textbf{99.88 $\pm$ 0.01} & \textbf{99.99 $\pm$ 0.00} \\
\arrayrulecolor{black!25}\midrule[0.3pt]\arrayrulecolor{black}
LightGBM & 99.66 $\pm$ 0.01 & 0.996 $\pm$ 0.000 & 99.98 $\pm$ 0.01 & 0.13 $\pm$ 0.00 & 99.85 $\pm$ 0.01 & 99.99 $\pm$ 0.00 \\
\arrayrulecolor{black!25}\midrule[0.3pt]\arrayrulecolor{black}
Random forest & 99.59 $\pm$ 0.01 & 0.995 $\pm$ 0.000 & 99.97 $\pm$ 0.00 & \textbf{0.08 $\pm$ 0.01} & 99.70 $\pm$ 0.01 & 99.98 $\pm$ 0.00 \\
\bottomrule
\end{tabularx}
\end{table}

\begin{table}[!htbp]
\caption{Binary detection on the NF-UNSW-NB15-v2 test partitions (\%, MCC without unit; mean $\pm$ SD over three seeds). Recall at 0.1\% or 1\% FPR is the highest recall at a false-positive rate not above that value; bold marks the best mean. E-GraphSAGE uses a host graph and is listed separately. The $k$NN-vote score takes only 11 values, so a recall at a fixed false-positive rate may not exist and is then 0.}
\label{tab:bin_nf}
\tablesize{\small}\renewcommand{\arraystretch}{1.3}
\setlength{\tabcolsep}{3pt}
\begin{tabularx}{\textwidth}{lCCCCCC}
\toprule
\textbf{Model} & \textbf{F1} & \textbf{MCC} & \textbf{PR-AUC} & \textbf{FPR} & \textbf{Recall at 0.1\% FPR} & \textbf{Recall at 1\% FPR} \\
\midrule
R-GCN & 90.57 $\pm$ 0.38 & 0.904 $\pm$ 0.004 & 95.11 $\pm$ 0.14 & 1.22 $\pm$ 0.06 & 43.62 $\pm$ 1.25 & 97.97 $\pm$ 0.35 \\
\arrayrulecolor{black!25}\midrule[0.3pt]\arrayrulecolor{black}
GCN & 90.82 $\pm$ 0.41 & 0.906 $\pm$ 0.004 & 95.04 $\pm$ 0.22 & 1.19 $\pm$ 0.05 & 44.33 $\pm$ 0.24 & 98.27 $\pm$ 0.21 \\
\arrayrulecolor{black!25}\midrule[0.3pt]\arrayrulecolor{black}
GraphSAGE & 90.87 $\pm$ 0.16 & 0.907 $\pm$ 0.002 & 94.82 $\pm$ 0.24 & 1.19 $\pm$ 0.02 & 42.31 $\pm$ 3.19 & 98.27 $\pm$ 0.46 \\
\arrayrulecolor{black!25}\midrule[0.3pt]\arrayrulecolor{black}
GAT & 90.96 $\pm$ 0.22 & 0.908 $\pm$ 0.002 & 95.01 $\pm$ 0.06 & 1.17 $\pm$ 0.03 & 43.53 $\pm$ 1.39 & 98.17 $\pm$ 0.21 \\
\arrayrulecolor{black!25}\midrule[0.3pt]\arrayrulecolor{black}
$k$NN vote & 92.47 $\pm$ 0.14 & 0.920 $\pm$ 0.001 & 94.95 $\pm$ 0.10 & 0.64 $\pm$ 0.02 & 0.00 $\pm$ 0.00 & 98.42 $\pm$ 0.03 \\
\arrayrulecolor{black!25}\midrule[0.3pt]\arrayrulecolor{black}
MLP & 90.57 $\pm$ 0.50 & 0.904 $\pm$ 0.005 & 94.90 $\pm$ 0.17 & 1.23 $\pm$ 0.07 & 40.79 $\pm$ 1.19 & 98.42 $\pm$ 0.23 \\
\arrayrulecolor{black!25}\midrule[0.3pt]\arrayrulecolor{black}
XGBoost & 91.92 $\pm$ 0.20 & 0.917 $\pm$ 0.002 & \textbf{96.94 $\pm$ 0.13} & 1.04 $\pm$ 0.03 & \textbf{56.95 $\pm$ 3.09} & \textbf{99.66 $\pm$ 0.15} \\
\arrayrulecolor{black!25}\midrule[0.3pt]\arrayrulecolor{black}
LightGBM & 91.79 $\pm$ 0.23 & 0.916 $\pm$ 0.002 & 96.81 $\pm$ 0.05 & 1.06 $\pm$ 0.03 & 55.72 $\pm$ 3.04 & 99.63 $\pm$ 0.13 \\
\arrayrulecolor{black!25}\midrule[0.3pt]\arrayrulecolor{black}
Random forest & \textbf{93.38 $\pm$ 0.18} & \textbf{0.930 $\pm$ 0.002} & 96.42 $\pm$ 0.13 & \textbf{0.51 $\pm$ 0.02} & 50.66 $\pm$ 4.37 & 99.34 $\pm$ 0.12 \\
\midrule
\multicolumn{7}{l}{\textit{Host-graph model}} \\
E-GraphSAGE & 83.12 $\pm$ 0.33 & 0.833 $\pm$ 0.003 & 78.23 $\pm$ 2.30 & 2.43 $\pm$ 0.06 & 8.42 $\pm$ 2.44 & 55.28 $\pm$ 3.89 \\
\bottomrule
\end{tabularx}
\end{table}

\begin{figure}[!htbp]
\centering
\includegraphics[width=\textwidth]{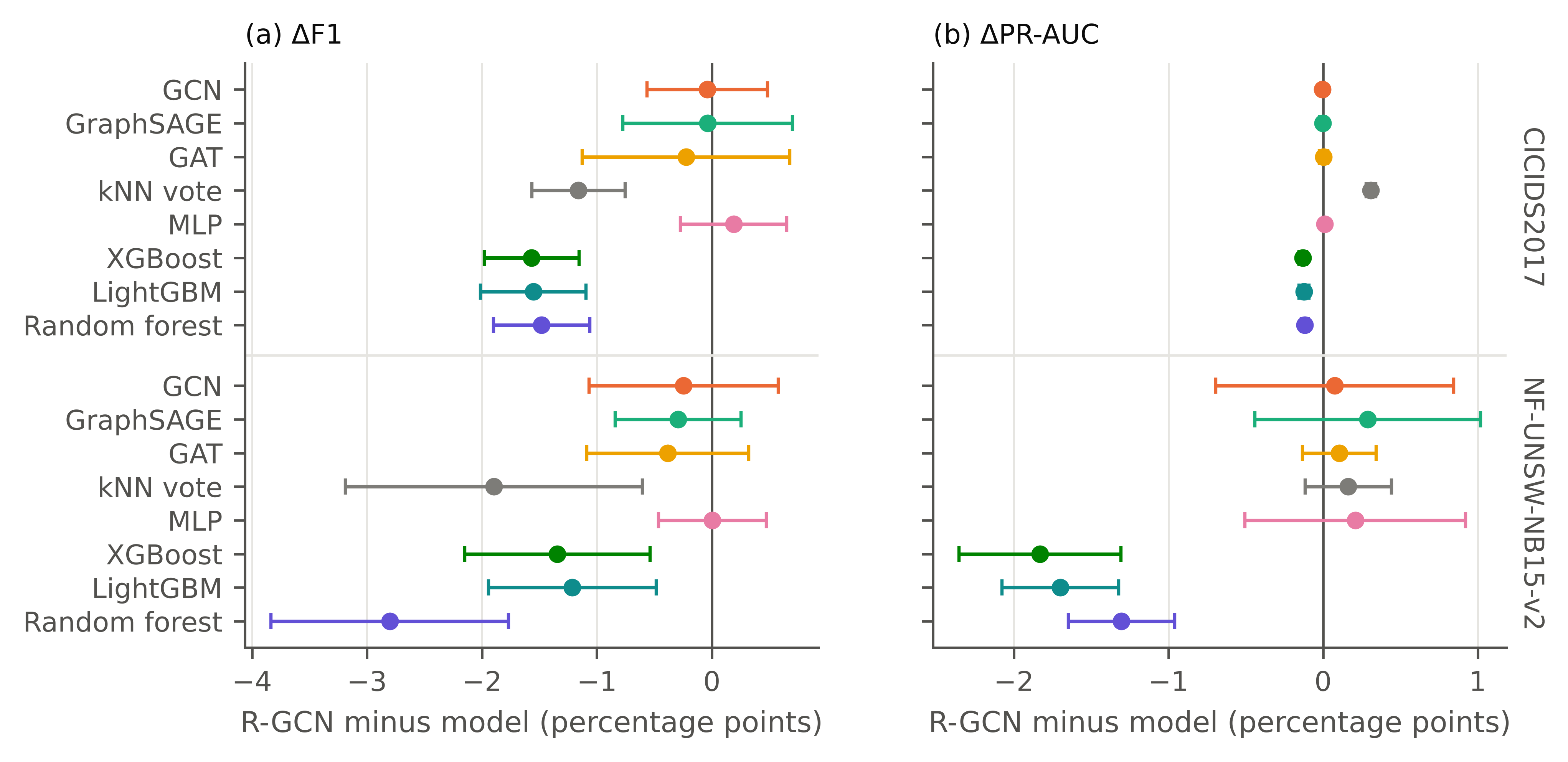}
\caption{Paired per-seed differences in binary detection, R-GCN minus each model, with 95\% confidence intervals from the
$t$ distribution (three seeds). (a) F1; (b) PR-AUC. Points left of zero favour the other model.}
\label{fig:forest}
\end{figure}

The paired differences in Table~\ref{tab:paired} separate these effects from seed-to-seed variation. Against GCN,
GraphSAGE, GAT and the MLP, the R-GCN's F1 differed by \textminus{}0.22 to +0.19 points on CICIDS2017 and by \textminus{}0.38 to 0.00
points on NF-UNSW-NB15-v2, and every interval contained zero. Against the three tree ensembles on CICIDS2017, the R-GCN was
1.48 to 1.57 points lower in F1 (Holm $p=0.030$ for each) and 0.12 to 0.13 points lower in PR-AUC. On NF-UNSW-NB15-v2 its
PR-AUC was 1.30 to 1.83 points lower than that of the ensembles (Holm $p$ between 0.021 and 0.026); its F1 was 1.21 to 2.80
points lower, with intervals that exclude zero but Holm-corrected $p$-values between 0.058 and 0.132.
Figure~\ref{fig:forest} plots the same intervals. The graph networks and the MLP cluster around zero on both datasets,
while every tree ensemble sits to the left of zero.

\begin{table}[!htbp]
\caption{Paired per-seed differences in binary detection, R-GCN minus each model (percentage points, three seeds), with 95\% confidence intervals from the $t$ distribution and paired $t$-test $p$-values Holm-corrected within each dataset and metric.}
\label{tab:paired}
\tablesize{\small}\renewcommand{\arraystretch}{1.3}
\setlength{\tabcolsep}{3pt}
\begin{tabularx}{\textwidth}{lCCcCCc}
\toprule
\textbf{R-GCN minus} & \textbf{$\Delta$F1} & \textbf{95\% CI} & \textbf{Holm $p$} & \textbf{$\Delta$PR-AUC} & \textbf{95\% CI} & \textbf{Holm $p$} \\
\midrule
\multicolumn{7}{l}{\textit{CICIDS2017}} \\
GCN & \textminus{}0.04 & [\textminus{}0.56, +0.48] & 1.000 & 0.00 & [\textminus{}0.02, +0.01] & 0.595 \\
\arrayrulecolor{black!25}\midrule[0.3pt]\arrayrulecolor{black}
GraphSAGE & \textminus{}0.04 & [\textminus{}0.77, +0.70] & 1.000 & 0.00 & [\textminus{}0.02, +0.02] & 1.000 \\
\arrayrulecolor{black!25}\midrule[0.3pt]\arrayrulecolor{black}
GAT & \textminus{}0.22 & [\textminus{}1.13, +0.68] & 1.000 & 0.00 & [\textminus{}0.02, +0.03] & 1.000 \\
\arrayrulecolor{black!25}\midrule[0.3pt]\arrayrulecolor{black}
$k$NN vote & \textminus{}1.16 & [\textminus{}1.57, \textminus{}0.75] & 0.033 & +0.31 & [+0.28, +0.34] & 0.004 \\
\arrayrulecolor{black!25}\midrule[0.3pt]\arrayrulecolor{black}
MLP & +0.19 & [\textminus{}0.27, +0.65] & 0.887 & +0.01 & [+0.01, +0.02] & 0.061 \\
\arrayrulecolor{black!25}\midrule[0.3pt]\arrayrulecolor{black}
XGBoost & \textminus{}1.57 & [\textminus{}1.98, \textminus{}1.16] & 0.030 & \textminus{}0.13 & [\textminus{}0.15, \textminus{}0.11] & 0.013 \\
\arrayrulecolor{black!25}\midrule[0.3pt]\arrayrulecolor{black}
LightGBM & \textminus{}1.55 & [\textminus{}2.01, \textminus{}1.09] & 0.030 & \textminus{}0.12 & [\textminus{}0.15, \textminus{}0.09] & 0.018 \\
\arrayrulecolor{black!25}\midrule[0.3pt]\arrayrulecolor{black}
Random forest & \textminus{}1.48 & [\textminus{}1.90, \textminus{}1.06] & 0.030 & \textminus{}0.12 & [\textminus{}0.14, \textminus{}0.09] & 0.013 \\
\midrule
\multicolumn{7}{l}{\textit{NF-UNSW-NB15-v2}} \\
GCN & \textminus{}0.25 & [\textminus{}1.07, +0.58] & 0.654 & +0.08 & [\textminus{}0.69, +0.84] & 0.786 \\
\arrayrulecolor{black!25}\midrule[0.3pt]\arrayrulecolor{black}
GraphSAGE & \textminus{}0.29 & [\textminus{}0.84, +0.25] & 0.574 & +0.29 & [\textminus{}0.44, +1.02] & 0.786 \\
\arrayrulecolor{black!25}\midrule[0.3pt]\arrayrulecolor{black}
GAT & \textminus{}0.38 & [\textminus{}1.09, +0.32] & 0.574 & +0.11 & [\textminus{}0.13, +0.34] & 0.786 \\
\arrayrulecolor{black!25}\midrule[0.3pt]\arrayrulecolor{black}
$k$NN vote & \textminus{}1.90 & [\textminus{}3.19, \textminus{}0.61] & 0.132 & +0.16 & [\textminus{}0.12, +0.44] & 0.647 \\
\arrayrulecolor{black!25}\midrule[0.3pt]\arrayrulecolor{black}
MLP & 0.00 & [\textminus{}0.47, +0.47] & 0.972 & +0.21 & [\textminus{}0.50, +0.92] & 0.786 \\
\arrayrulecolor{black!25}\midrule[0.3pt]\arrayrulecolor{black}
XGBoost & \textminus{}1.34 & [\textminus{}2.15, \textminus{}0.54] & 0.132 & \textminus{}1.83 & [\textminus{}2.35, \textminus{}1.31] & 0.026 \\
\arrayrulecolor{black!25}\midrule[0.3pt]\arrayrulecolor{black}
LightGBM & \textminus{}1.21 & [\textminus{}1.94, \textminus{}0.48] & 0.132 & \textminus{}1.70 & [\textminus{}2.08, \textminus{}1.32] & 0.021 \\
\arrayrulecolor{black!25}\midrule[0.3pt]\arrayrulecolor{black}
Random forest & \textminus{}2.80 & [\textminus{}3.83, \textminus{}1.77] & 0.058 & \textminus{}1.30 & [\textminus{}1.65, \textminus{}0.96] & 0.026 \\
\bottomrule
\end{tabularx}
\end{table}

\subsection{Multiclass detection}
The multiclass task separates the model families more sharply (Tables~\ref{tab:multi_cic} and~\ref{tab:multi_nf}). On CICIDS2017 the weighted F1 of
every model exceeded 96\%, but macro-F1, which gives each class equal weight, was 64.51\% for the R-GCN and 64.65\% for the
GCN, against 93.08\% for XGBoost, 88.37\% for the random forest and 81.52\% for the $k$NN vote. On NF-UNSW-NB15-v2 macro-F1
ranged from 44.21\% (R-GCN) to 55.98\% (XGBoost). The gap between weighted and macro-F1 reflects the small attack classes,
which carry the same weight as Benign in macro-F1.

\begin{table}[!htbp]
\caption{Multiclass detection on the CICIDS2017 test partitions (\%, mean $\pm$ SD over three seeds). Macro-F1 averages the per-class F1 over the classes present in the test partition; weighted F1 weights each class by its number of flows; bold marks the best mean.}
\label{tab:multi_cic}
\tablesize{\small}\renewcommand{\arraystretch}{1.3}
\setlength{\tabcolsep}{3pt}
\begin{tabularx}{\textwidth}{lCCC}
\toprule
\textbf{Model} & \textbf{Macro-F1} & \textbf{Weighted F1} & \textbf{Accuracy} \\
\midrule
R-GCN & 64.51 $\pm$ 0.23 & 97.80 $\pm$ 0.05 & 96.60 $\pm$ 0.04 \\
\arrayrulecolor{black!25}\midrule[0.3pt]\arrayrulecolor{black}
GCN & 64.65 $\pm$ 0.38 & 97.85 $\pm$ 0.15 & 96.65 $\pm$ 0.21 \\
\arrayrulecolor{black!25}\midrule[0.3pt]\arrayrulecolor{black}
$k$NN vote & 81.52 $\pm$ 1.81 & 99.74 $\pm$ 0.01 & 99.74 $\pm$ 0.01 \\
\arrayrulecolor{black!25}\midrule[0.3pt]\arrayrulecolor{black}
MLP & 62.39 $\pm$ 0.59 & 96.47 $\pm$ 0.71 & 94.65 $\pm$ 1.09 \\
\arrayrulecolor{black!25}\midrule[0.3pt]\arrayrulecolor{black}
XGBoost & \textbf{93.08 $\pm$ 1.22} & 99.85 $\pm$ 0.01 & 99.84 $\pm$ 0.01 \\
\arrayrulecolor{black!25}\midrule[0.3pt]\arrayrulecolor{black}
LightGBM & 81.57 $\pm$ 0.75 & 99.66 $\pm$ 0.01 & 99.58 $\pm$ 0.04 \\
\arrayrulecolor{black!25}\midrule[0.3pt]\arrayrulecolor{black}
Random forest & 88.37 $\pm$ 2.11 & \textbf{99.85 $\pm$ 0.01} & \textbf{99.85 $\pm$ 0.01} \\
\bottomrule
\end{tabularx}
\end{table}

\begin{table}[!htbp]
\caption{Multiclass detection on the NF-UNSW-NB15-v2 test partitions (\%, mean $\pm$ SD over three seeds). Macro-F1 averages the per-class F1 over the classes present in the test partition; weighted F1 weights each class by its number of flows; bold marks the best mean.}
\label{tab:multi_nf}
\tablesize{\small}\renewcommand{\arraystretch}{1.3}
\setlength{\tabcolsep}{3pt}
\begin{tabularx}{\textwidth}{lCCC}
\toprule
\textbf{Model} & \textbf{Macro-F1} & \textbf{Weighted F1} & \textbf{Accuracy} \\
\midrule
R-GCN & 44.21 $\pm$ 1.07 & 97.04 $\pm$ 0.11 & 96.25 $\pm$ 0.21 \\
\arrayrulecolor{black!25}\midrule[0.3pt]\arrayrulecolor{black}
$k$NN vote & 48.72 $\pm$ 0.31 & 97.79 $\pm$ 0.01 & \textbf{97.88 $\pm$ 0.01} \\
\arrayrulecolor{black!25}\midrule[0.3pt]\arrayrulecolor{black}
MLP & 45.85 $\pm$ 1.24 & 97.02 $\pm$ 0.17 & 96.19 $\pm$ 0.33 \\
\arrayrulecolor{black!25}\midrule[0.3pt]\arrayrulecolor{black}
XGBoost & \textbf{55.98 $\pm$ 0.88} & \textbf{97.90 $\pm$ 0.06} & 97.48 $\pm$ 0.08 \\
\arrayrulecolor{black!25}\midrule[0.3pt]\arrayrulecolor{black}
LightGBM & 53.82 $\pm$ 1.04 & 97.83 $\pm$ 0.05 & 97.39 $\pm$ 0.07 \\
\bottomrule
\end{tabularx}
\end{table}

\begin{figure}[!htbp]
\centering
\includegraphics[width=\textwidth]{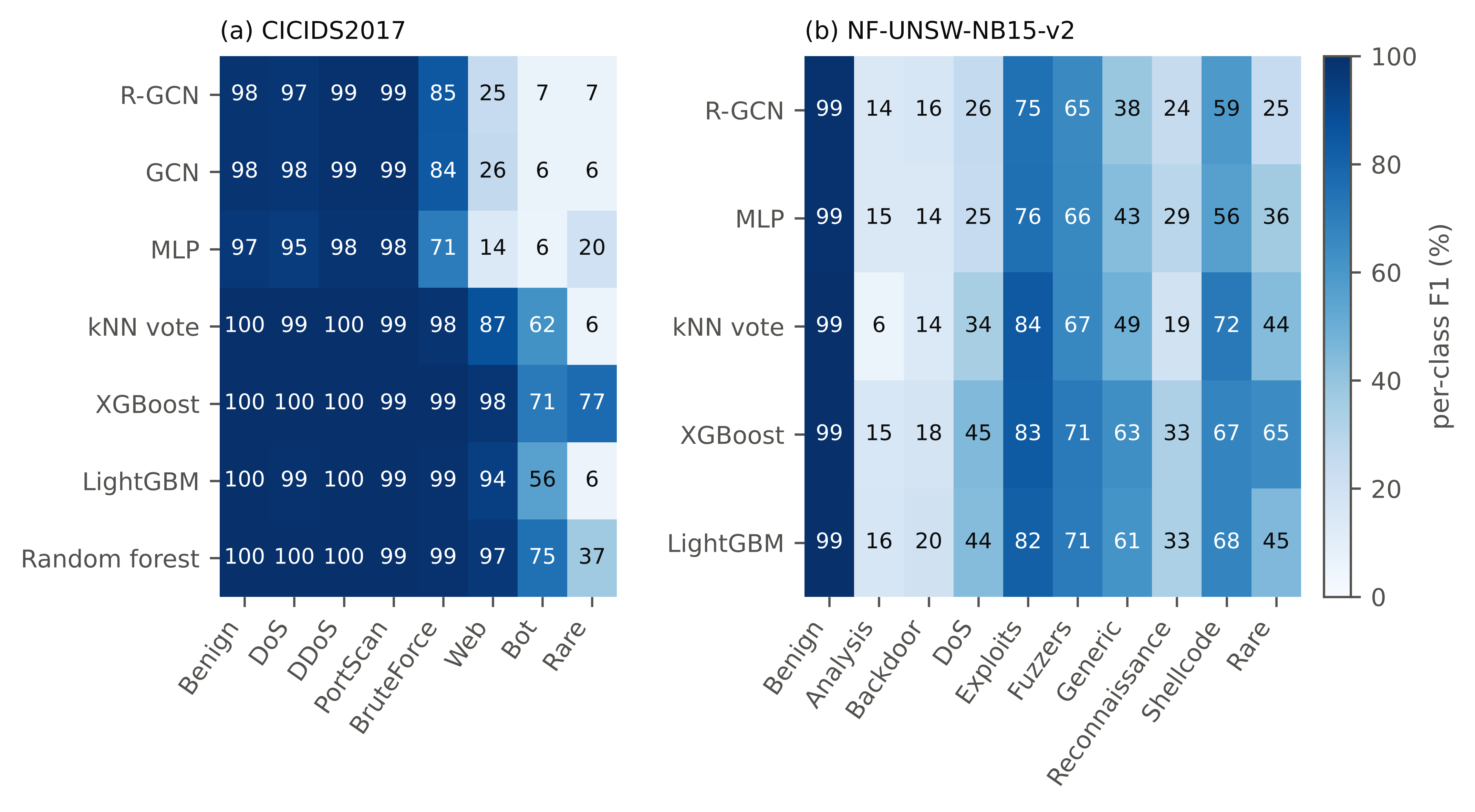}
\caption{Per-class F1 in multiclass detection on the test partitions (\%, mean over three seeds). (a) CICIDS2017; (b)
NF-UNSW-NB15-v2. Classes follow the order of Tables~\ref{tab:data_cic} and~\ref{tab:data_nf}.}
\label{fig:perclass}
\end{figure}

Figure~\ref{fig:perclass} breaks macro-F1 down by class. On CICIDS2017 the four large classes (Benign, DoS, DDoS and
PortScan) reached a per-class F1 of at least 95\% for every model, so the differences arise in the small classes. For Web
attacks the R-GCN reached 24.83\%, against 87.09\% for the $k$NN vote and 97.64\% for XGBoost; for Bot the corresponding
values were 6.51\%, 62.47\% and 71.30\%. On NF-UNSW-NB15-v2 no model exceeded 20\% F1 on Analysis or Backdoor flows, and
the tree ensembles led mainly on DoS, Generic and the Rare class.

\begin{figure}[!htbp]
\centering
\includegraphics[width=\textwidth]{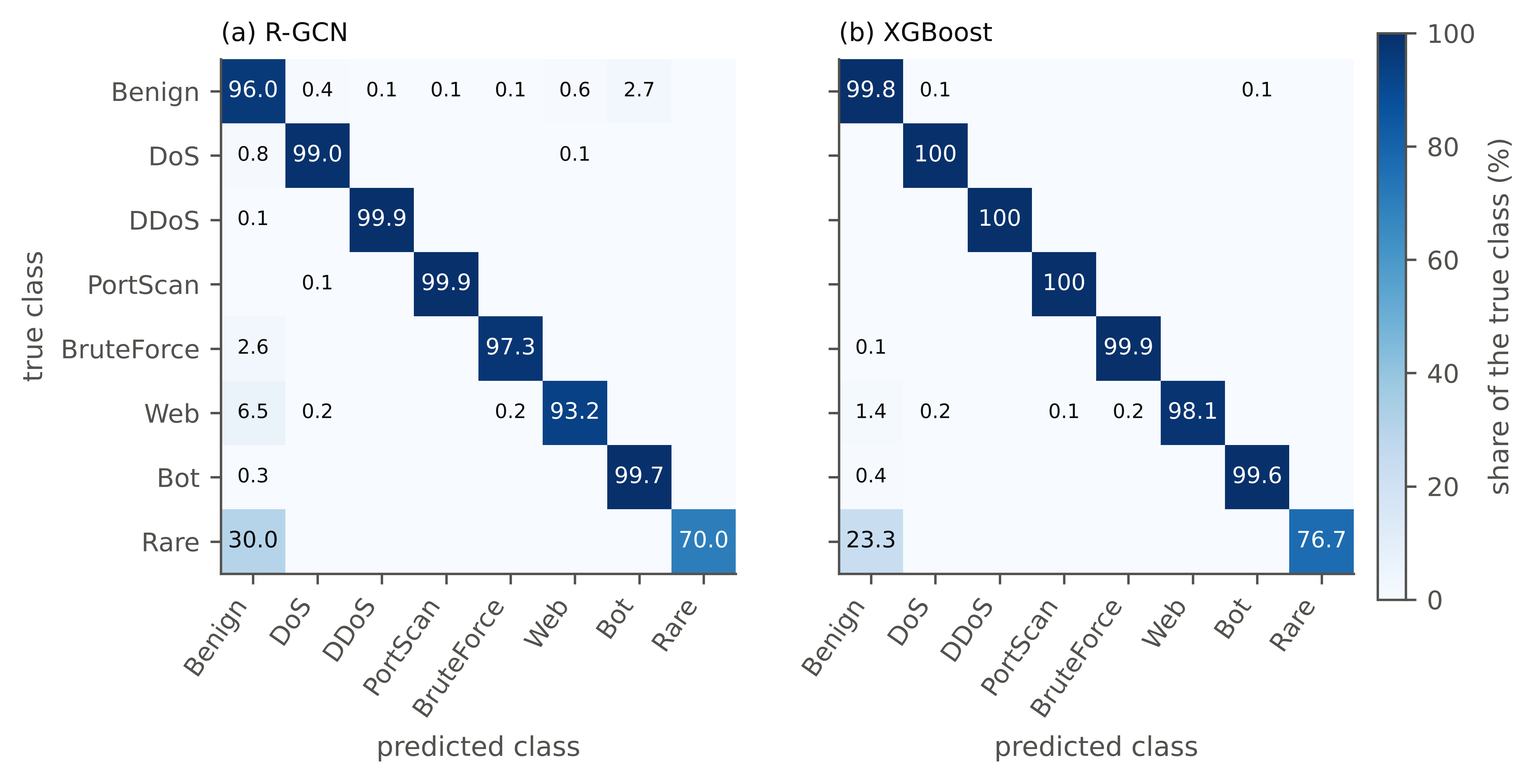}
\caption{Confusion matrices of multiclass detection on CICIDS2017, summed over the three seeds and normalised by the true
class (rows, \%). (a) R-GCN; (b) XGBoost. Cells below 0.05\% are left blank.}
\label{fig:confusion}
\end{figure}

The confusion matrices in Figure~\ref{fig:confusion} show that the low scores of the R-GCN on these classes come from false
alarms rather than missed attacks. The R-GCN recalled 93.16\% of Web flows and 99.66\% of Bot flows, close to XGBoost
(98.06\% and 99.57\%). Its precision, however, was 14.32\% for Web and 3.36\% for Bot, against 97.22\% and 55.49\% for
XGBoost. Over the three seeds the R-GCN labelled 33,402 of 1,257,034 benign test flows as Bot and 7,026 as Web; XGBoost
labelled 932 and 23. Because Benign is by far the largest class, a share of benign flows that is small in percentage terms
outnumbers the true members of a small attack class many times over.

\subsection{Ablation}
\begin{figure}[!htbp]
\centering
\includegraphics[width=\textwidth]{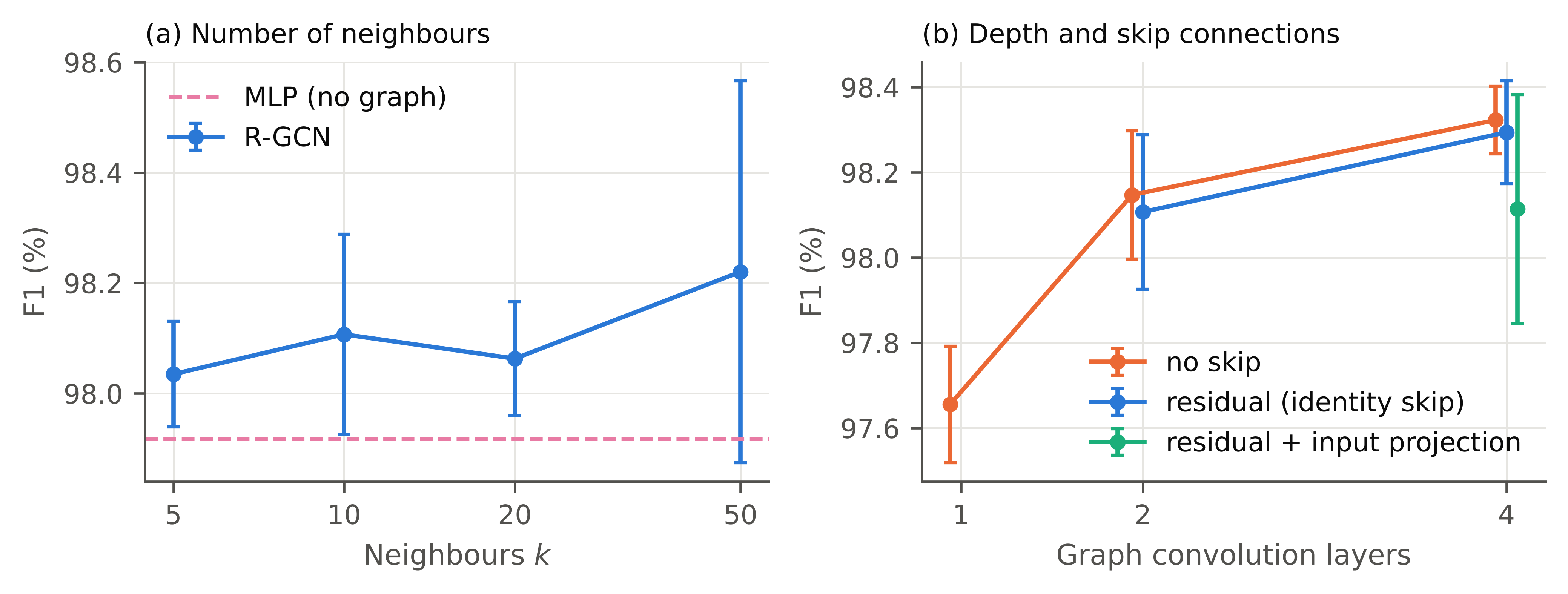}
\caption{Ablation on CICIDS2017, binary F1 (mean $\pm$ SD over three seeds). (a) Number of neighbours $k$ (logarithmic
axis); the dashed line is the MLP without a graph. (b) Number of graph convolution layers without a skip, with the identity
skip of the R-GCN, and with an additional input projection.}
\label{fig:kdepth}
\end{figure}

Table~\ref{tab:abl} and Figure~\ref{fig:kdepth} change one setting of the R-GCN at a time on CICIDS2017. The neighbourhood
size moved F1 within 98.03\% to 98.22\% for $k$ between 5 and 50. Removing the residual connection at two layers changed F1
by +0.04 points, and at four layers the networks with and without the skip reached 98.29\% and 98.32\%. Building the
graph transductively, over training, validation and test features, changed F1 by \textminus{}0.11 points (interval \textminus{}0.29 to
+0.07), and adding the destination port changed it by +0.51 points (interval \textminus{}0.47 to +1.48). The unweighted
cross-entropy was the only setting whose interval excluded zero (+0.72 points, interval +0.18 to +1.27), mainly
through a lower false-positive rate (0.16\% against 0.64\%); after Holm correction over the table its $p$-value was 0.382.

\begin{table}[!htbp]
\caption{Ablation of the R-GCN on CICIDS2017, binary detection, one setting changed at a time (\%, mean $\pm$ SD over three seeds). $\Delta$F1 is the paired difference to the reference in percentage points with its 95\% confidence interval; Holm $p$ is corrected over all rows.}
\label{tab:abl}
\tablesize{\small}\renewcommand{\arraystretch}{1.3}
\setlength{\tabcolsep}{3pt}
\begin{tabularx}{\textwidth}{>{\hsize=1.45\hsize}L>{\hsize=0.75\hsize}C>{\hsize=0.75\hsize}C>{\hsize=0.7\hsize}C>{\hsize=1.35\hsize}Cc}
\toprule
\textbf{Configuration} & \textbf{F1} & \textbf{PR-AUC} & \textbf{FPR} & \textbf{$\Delta$F1 [95\% CI]} & \textbf{Holm $p$} \\
\midrule
R-GCN reference (2 layers, residual, $k=10$) & 98.11 $\pm$ 0.18 & 99.86 $\pm$ 0.01 & 0.64 $\pm$ 0.07 & -- & -- \\
\arrayrulecolor{black!25}\midrule[0.3pt]\arrayrulecolor{black}
2 layers, no skip (GCN) & 98.15 $\pm$ 0.15 & 99.86 $\pm$ 0.01 & 0.62 $\pm$ 0.06 & +0.04 [\textminus{}0.48, +0.56] & 1.000 \\
\arrayrulecolor{black!25}\midrule[0.3pt]\arrayrulecolor{black}
$k=5$ & 98.03 $\pm$ 0.10 & 99.85 $\pm$ 0.01 & 0.67 $\pm$ 0.03 & \textminus{}0.07 [\textminus{}0.30, +0.16] & 1.000 \\
\arrayrulecolor{black!25}\midrule[0.3pt]\arrayrulecolor{black}
$k=20$ & 98.06 $\pm$ 0.10 & 99.84 $\pm$ 0.01 & 0.66 $\pm$ 0.04 & \textminus{}0.04 [\textminus{}0.27, +0.18] & 1.000 \\
\arrayrulecolor{black!25}\midrule[0.3pt]\arrayrulecolor{black}
$k=50$ & 98.22 $\pm$ 0.35 & 99.82 $\pm$ 0.01 & 0.53 $\pm$ 0.20 & +0.11 [\textminus{}0.30, +0.53] & 1.000 \\
\arrayrulecolor{black!25}\midrule[0.3pt]\arrayrulecolor{black}
1 layer & 97.66 $\pm$ 0.14 & 99.75 $\pm$ 0.02 & 0.75 $\pm$ 0.13 & \textminus{}0.45 [\textminus{}1.24, +0.34] & 1.000 \\
\arrayrulecolor{black!25}\midrule[0.3pt]\arrayrulecolor{black}
4 layers, no skip & 98.32 $\pm$ 0.08 & 99.88 $\pm$ 0.01 & 0.56 $\pm$ 0.03 & +0.22 [\textminus{}0.25, +0.68] & 1.000 \\
\arrayrulecolor{black!25}\midrule[0.3pt]\arrayrulecolor{black}
4 layers, residual & 98.29 $\pm$ 0.12 & 99.87 $\pm$ 0.01 & 0.57 $\pm$ 0.05 & +0.19 [\textminus{}0.12, +0.49] & 1.000 \\
\arrayrulecolor{black!25}\midrule[0.3pt]\arrayrulecolor{black}
4 layers, residual + input projection & 98.11 $\pm$ 0.27 & 99.83 $\pm$ 0.01 & 0.60 $\pm$ 0.17 & +0.01 [\textminus{}1.09, +1.10] & 1.000 \\
\arrayrulecolor{black!25}\midrule[0.3pt]\arrayrulecolor{black}
unweighted cross-entropy & 98.83 $\pm$ 0.04 & 99.85 $\pm$ 0.01 & 0.16 $\pm$ 0.02 & +0.72 [+0.18, +1.27] & 0.382 \\
\arrayrulecolor{black!25}\midrule[0.3pt]\arrayrulecolor{black}
focal loss ($\gamma=2$) & 98.11 $\pm$ 0.34 & 99.83 $\pm$ 0.01 & 0.60 $\pm$ 0.19 & 0.00 [\textminus{}0.40, +0.40] & 1.000 \\
\arrayrulecolor{black!25}\midrule[0.3pt]\arrayrulecolor{black}
no batch normalisation & 98.41 $\pm$ 0.39 & 99.85 $\pm$ 0.01 & 0.45 $\pm$ 0.21 & +0.30 [\textminus{}0.37, +0.98] & 1.000 \\
\arrayrulecolor{black!25}\midrule[0.3pt]\arrayrulecolor{black}
destination port added & 98.61 $\pm$ 0.24 & 99.93 $\pm$ 0.01 & 0.47 $\pm$ 0.16 & +0.51 [\textminus{}0.47, +1.48] & 1.000 \\
\arrayrulecolor{black!25}\midrule[0.3pt]\arrayrulecolor{black}
transductive graph construction & 98.00 $\pm$ 0.12 & 99.86 $\pm$ 0.01 & 0.69 $\pm$ 0.04 & \textminus{}0.11 [\textminus{}0.29, +0.07] & 1.000 \\
\bottomrule
\end{tabularx}
\end{table}

\subsection{Unseen attack families and feature noise}
\begin{figure}[!htbp]
\centering
\includegraphics[width=\textwidth]{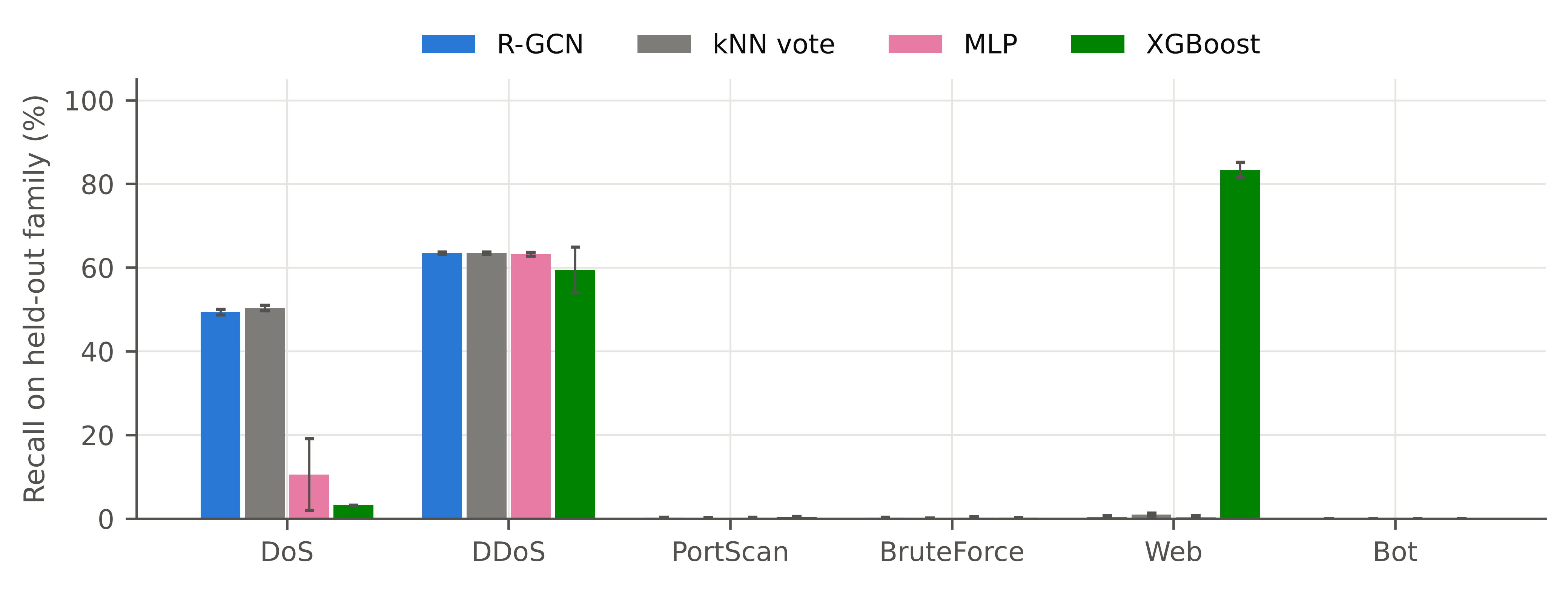}
\caption{Recall on the attack family removed from training and validation (CICIDS2017, threshold 0.5, mean $\pm$ SD over
three seeds).}
\label{fig:lofo}
\end{figure}

When a family was absent from training and validation, detection depended on whether its flows resembled a family that was
present (Table~\ref{tab:lofo}, Figure~\ref{fig:lofo}). Held-out DDoS flows were detected at 59.42\% to 63.42\% by all four
models. Held-out DoS flows were detected at 49.37\% by the R-GCN and 50.36\% by the $k$NN vote, but only at 10.49\% by the
MLP and 3.19\% by XGBoost. Held-out Web attacks were detected at 83.36\% by XGBoost and at no more than 1.01\% by the other
models. PortScan, BruteForce and Bot flows were almost never detected (recall at most 0.39\%). On the families seen in
training, recall stayed between 99.26\% and 99.95\%.

\begin{table}[!htbp]
\caption{Recall (\%) on the attack family withheld from training and validation, CICIDS2017, threshold 0.5 (mean $\pm$ SD over three seeds). Flows: test flows of the withheld family; the last two rows summarise the families seen in training and the benign flows over all six runs.}
\label{tab:lofo}
\tablesize{\small}\renewcommand{\arraystretch}{1.3}
\setlength{\tabcolsep}{3pt}
\begin{tabularx}{\textwidth}{lrCCCC}
\toprule
\textbf{Held-out family} & \textbf{Flows} & \textbf{R-GCN} & \textbf{$k$NN vote} & \textbf{MLP} & \textbf{XGBoost} \\
\midrule
DoS & 38,749 & 49.37 $\pm$ 0.67 & \textbf{50.36 $\pm$ 0.68} & 10.49 $\pm$ 8.56 & 3.19 $\pm$ 0.07 \\
\arrayrulecolor{black!25}\midrule[0.3pt]\arrayrulecolor{black}
DDoS & 25,603 & 63.42 $\pm$ 0.27 & \textbf{63.42 $\pm$ 0.27} & 63.16 $\pm$ 0.44 & 59.42 $\pm$ 5.47 \\
\arrayrulecolor{black!25}\midrule[0.3pt]\arrayrulecolor{black}
PortScan & 18,139 & 0.24 $\pm$ 0.08 & 0.18 $\pm$ 0.03 & 0.28 $\pm$ 0.10 & \textbf{0.39 $\pm$ 0.11} \\
\arrayrulecolor{black!25}\midrule[0.3pt]\arrayrulecolor{black}
BruteForce & 1,830 & 0.24 $\pm$ 0.14 & 0.09 $\pm$ 0.08 & 0.24 $\pm$ 0.17 & \textbf{0.26 $\pm$ 0.03} \\
\arrayrulecolor{black!25}\midrule[0.3pt]\arrayrulecolor{black}
Web & 428 & 0.31 $\pm$ 0.36 & 1.01 $\pm$ 0.36 & 0.31 $\pm$ 0.36 & \textbf{83.36 $\pm$ 1.81} \\
\arrayrulecolor{black!25}\midrule[0.3pt]\arrayrulecolor{black}
Bot & 389 & \textbf{0.00 $\pm$ 0.00} & \textbf{0.00 $\pm$ 0.00} & \textbf{0.00 $\pm$ 0.00} & \textbf{0.00 $\pm$ 0.00} \\
\arrayrulecolor{black!25}\midrule[0.3pt]\arrayrulecolor{black}
Seen families (recall) & -- & 99.32 $\pm$ 0.15 & 99.26 $\pm$ 0.18 & 99.29 $\pm$ 0.13 & 99.95 $\pm$ 0.02 \\
\arrayrulecolor{black!25}\midrule[0.3pt]\arrayrulecolor{black}
Benign flows (FPR) & -- & 0.57 $\pm$ 0.12 & 0.13 $\pm$ 0.02 & 0.58 $\pm$ 0.15 & 0.10 $\pm$ 0.02 \\
\bottomrule
\end{tabularx}
\end{table}

\begin{figure}[!htbp]
\centering
\includegraphics[width=0.6\textwidth]{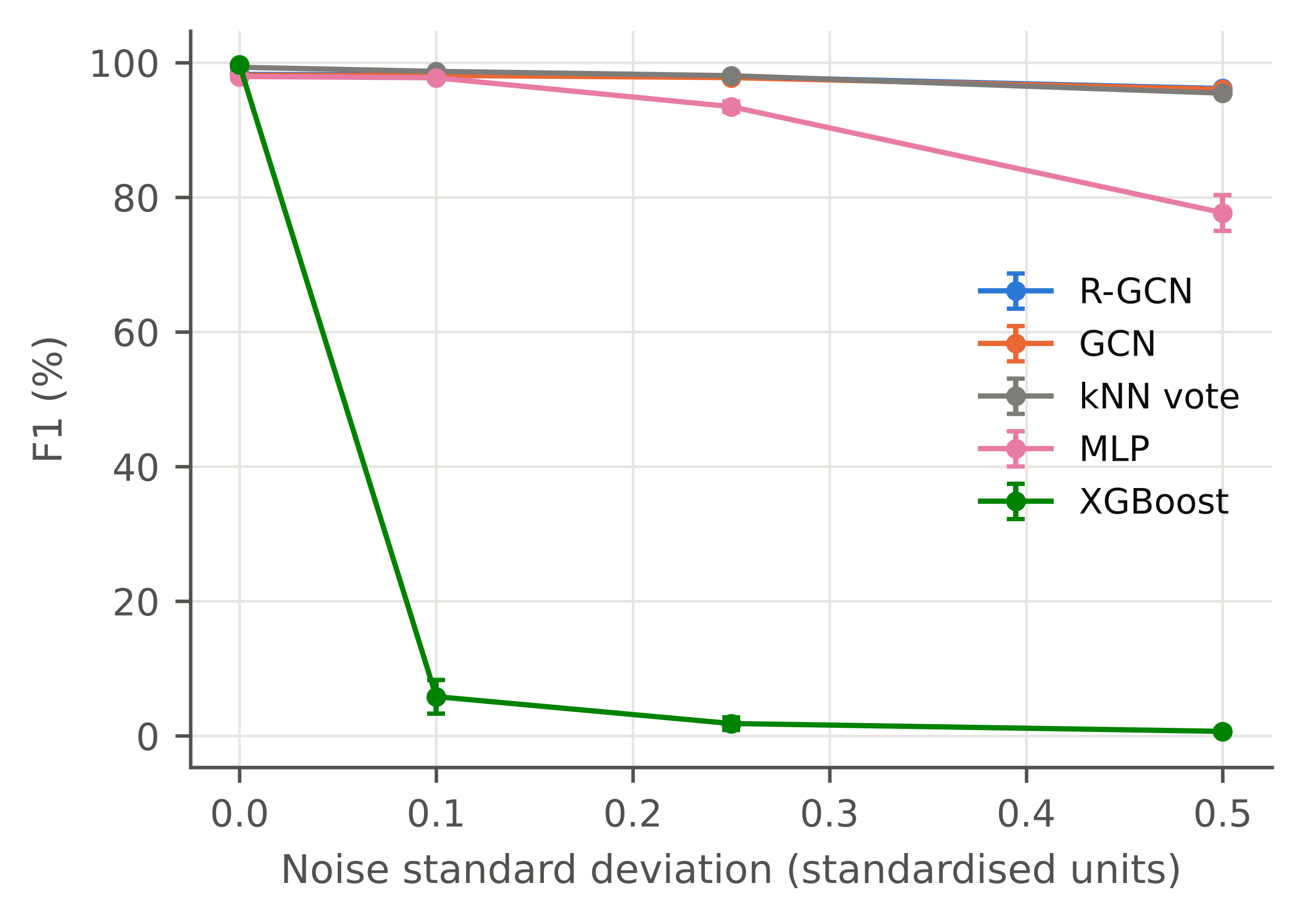}
\caption{Binary F1 under Gaussian noise added to the standardised test features (CICIDS2017, up to 100,000 test flows per
run, mean $\pm$ SD over three seeds).}
\label{fig:noise}
\end{figure}

Gaussian noise added to the standardised test features affected the model families very differently (Table~\ref{tab:noise},
Figure~\ref{fig:noise}). At a noise standard deviation of 0.5, the R-GCN kept an F1 of 96.16\% (98.26\% without noise on the
same test subset), the GCN 96.04\% and the $k$NN vote 95.47\%, while the MLP fell to 77.70\%. XGBoost fell to 5.84\% already
at a standard deviation of 0.1 and to 0.67\% at 0.5. At 0.1 its precision stayed at 95.95\% and its PR-AUC at 80.93\%, but
its recall dropped to 3.02\%: most attack flows were scored below the decision threshold.

\begin{table}[!htbp]
\caption{Binary F1 (\%) on CICIDS2017 under Gaussian noise with standard deviation $\sigma$ added to the standardised test features (up to 100,000 test flows; mean $\pm$ SD over three seeds).}
\label{tab:noise}
\tablesize{\small}\renewcommand{\arraystretch}{1.3}
\setlength{\tabcolsep}{3pt}
\begin{tabularx}{\textwidth}{lCCCC}
\toprule
\textbf{Model} & \textbf{$\sigma=0$} & \textbf{$\sigma=0.1$} & \textbf{$\sigma=0.25$} & \textbf{$\sigma=0.5$} \\
\midrule
R-GCN & 98.26 $\pm$ 0.24 & 98.21 $\pm$ 0.21 & 97.91 $\pm$ 0.25 & 96.16 $\pm$ 0.29 \\
\arrayrulecolor{black!25}\midrule[0.3pt]\arrayrulecolor{black}
GCN & 98.12 $\pm$ 0.07 & 98.11 $\pm$ 0.08 & 97.78 $\pm$ 0.05 & 96.04 $\pm$ 0.08 \\
\arrayrulecolor{black!25}\midrule[0.3pt]\arrayrulecolor{black}
$k$NN vote & 99.32 $\pm$ 0.03 & 98.71 $\pm$ 0.01 & 98.07 $\pm$ 0.07 & 95.47 $\pm$ 0.05 \\
\arrayrulecolor{black!25}\midrule[0.3pt]\arrayrulecolor{black}
MLP & 97.94 $\pm$ 0.04 & 97.76 $\pm$ 0.13 & 93.47 $\pm$ 0.78 & 77.70 $\pm$ 2.65 \\
\arrayrulecolor{black!25}\midrule[0.3pt]\arrayrulecolor{black}
XGBoost & 99.69 $\pm$ 0.04 & 5.84 $\pm$ 2.51 & 1.84 $\pm$ 0.96 & 0.67 $\pm$ 0.40 \\
\bottomrule
\end{tabularx}
\end{table}

\section{Discussion}
The central result is that message passing did not improve detection once the graph was built from training flows only.
In the binary task the R-GCN, GCN, GraphSAGE and GAT were indistinguishable from an MLP that ignores the graph on both
datasets, and their multiclass scores stayed close to those of the MLP. The residual connection had no measurable effect
at the two and four layers tested, so the loss of information in deep stacks of graph convolutions that motivates such
connections \cite{liq2018} played no visible role here. The $k$NN vote, which only reads the labels of the same neighbours,
reached a higher F1 than every graph network on both datasets. The useful signal in the neighbourhood lies largely in
its label composition. The learned aggregation did not add to it.

Tree ensembles were the strongest detectors, consistent with tabular comparisons in this domain \cite{zoppi2024}. Their
margin was small in binary F1 but large where minority classes matter. In macro-F1 on CICIDS2017, XGBoost exceeded the two
graph networks by more than 28 points. On NF-UNSW-NB15-v2 it led in recall at a false-positive rate of 0.1\%, the
operating region in which a deployed detector has to work. For a detector that must label known attack types on clean flow
features, a gradient-boosted ensemble is the better choice on this evidence.

The per-class results locate the multiclass gap precisely. The neural models found almost all Web and Bot flows but paid
for it with many benign flows labelled as those classes, so their precision on small classes collapsed while their recall
stayed high. A likely explanation is that class weights of up to 100 pull a smooth neural decision
boundary into the dense benign region next to each small class. The axis-aligned splits of a tree ensemble can isolate a
small class without moving the boundary elsewhere, and the boosted trees kept far higher precision with the same weights.
The binary ablation points the same way: the unweighted cross-entropy raised the R-GCN's F1 by 0.72 points, mainly through
a lower false-positive rate. Weight caps, per-class thresholds chosen on the validation partition and a second-stage classifier
for flows assigned to small classes offer direct ways to reduce these false alarms.

The false-positive rates translate directly into analyst workload. At a threshold of 0.5 the R-GCN flagged 0.64\% of
benign CICIDS2017 test flows, about 640 false alarms per 100,000 benign flows, against about 80 to 130 for the tree
ensembles. Recall at a fixed false-positive rate compares the models at a budget the operator chooses: at 0.1\% the
ensembles kept 99.70\% to 99.88\% of attack flows on CICIDS2017 against 97.39\% for the R-GCN, and on NF-UNSW-NB15-v2
XGBoost kept 56.95\% against 43.62\%. Threshold-free scores understate this gap. On NF-UNSW-NB15-v2 the two models differed
by less than two points in PR-AUC (96.94\% against 95.11\%) but by more than 13 points in recall at the strict budget. No
model reached high recall at that budget on the second dataset, so a detector deployed on similar traffic would need a
looser false-positive budget or a second stage that re-examines flagged flows.

The host-graph comparison points the same way. E-GraphSAGE \cite{lo2022}, run on NF-UNSW-NB15-v2 under the shared
protocol, reached a binary F1 of 83.12\%, below every flow-level model in Table~\ref{tab:bin_nf}. Connecting flows through the
hosts they share, as in host graphs \cite{bilot2023}, did not compensate for the per-flow information that the
feature-based models use directly.

The transductive ablation matters for graph-based intrusion detection in general. Letting test features enter the graph
did not change F1 here, so the inductive restriction costs nothing in accuracy while removing the dependence of each
prediction on the other test flows. Models evaluated transductively, or with a separate graph built inside each test
partition, can be re-evaluated inductively in the same way, and reported gains that rely on test-time graph construction
deserve this check.

The graph-based models showed two strengths. Under feature noise they lost two to four points of F1 at the strongest
level tested, whereas XGBoost lost almost all of its recall at the smallest level. Its PR-AUC remained at 80.93\%, so the
noise mainly shifted its scores across the fixed threshold, which suggests that splits on precise feature values transfer
poorly to perturbed inputs. In the leave-one-family-out test the R-GCN and the $k$NN vote recovered about half of the
held-out DoS flows, against a few percent for the MLP and XGBoost, probably because these flows lie close to training
flows of other attack families in feature space. Both strengths rest on resemblance to known traffic: PortScan,
BruteForce and Bot flows were rarely detected when their family was withheld, and only XGBoost recognised withheld Web
attacks.

Label-aware de-duplication shaped the benchmarks themselves. Removing exact duplicates reduced NF-UNSW-NB15-v2 from
2,390,275 to 731,810 flows; without this step, identical flows would appear in training and test partitions and every model
would be rewarded for memorisation. With resampled splits, the standard deviation of binary F1 was at most 0.50 points for every
model, small enough to show that the graph networks do not differ from each other and to resolve their gap to the tree
ensembles on CICIDS2017.

Three practices follow for studies of graph-based detectors on flow data, in line with the call for evaluation closer to
deployment \cite{apruzzese2023}. First, a feature-only MLP and a $k$NN vote over the same neighbours belong next to any
graph model: together they separate the contribution of the learned aggregation from that of the features and of the
neighbours' labels. Second, the graph should be built from training flows only, with validation and test flows attached at
prediction time, and exact duplicates should be removed within each dataset before the split. Third, comparisons should
report every seed and the paired differences, since on CICIDS2017 the whole range of the five neural models, 97.92\% to
98.33\% in F1, was of the same order as their standard deviations across seeds.

\section{Conclusions}
A residual graph convolutional network on a nearest-neighbour graph of flows, evaluated with an inductive graph, label-aware
de-duplication and resampled splits, matched a multilayer perceptron and the other graph networks on CICIDS2017 and
NF-UNSW-NB15-v2, and it trailed gradient-boosted trees and random forests, most clearly in multiclass macro-F1, where
benign flows assigned to small attack classes lowered its precision. Its residual
connection, the neighbourhood size and transductive graph construction made no measurable difference. The graph-based
models were markedly more robust to feature noise than XGBoost and recovered part of an unseen DoS family, but no model
detected most unseen attack families. Evaluations of graph models for intrusion detection should therefore include a
same-neighbour vote, tree ensembles and an inductive graph before attributing gains to message passing.
Future work will train the detector across several organisations with federated learning \cite{iqbal2021fl}, where
differences in traffic and in model capacity between sites can be handled by knowledge distillation from heterogeneous
clients \cite{ma2021,iqbal2026}.Deployment on Internet of Things and smart-city networks is a further
direction, since security mechanisms there must fit the limited computing, energy and bandwidth budgets of the devices
\cite{ramachandran2026}.

\end{document}